\documentclass[conference]{IEEEtran}
\IEEEoverridecommandlockouts
\usepackage{amsmath,amsfonts}
\usepackage{algorithmic}
\usepackage[linesnumbered, ruled]{algorithm2e}
\usepackage{array}
\usepackage{textcomp}
\usepackage{stfloats}
\usepackage{url}
\usepackage{verbatim}
\usepackage{graphicx}
\usepackage{cite}
\usepackage{xspace}
\usepackage{enumitem}
\usepackage{booktabs}
\usepackage{multirow}
\usepackage{subfigure}
\usepackage{hyperref}
\def\BibTeX{{\rm B\kern-.05em{\sc i\kern-.025em b}\kern-.08em
    T\kern-.1667em\lower.7ex\hbox{E}\kern-.125emX}}
\newcommand{\xhdr}[1]{{\bfseries #1}.}
\newcommand{\name}{STRIPE\xspace}

\begin{document}

\title{Detecting Anomalies in Dynamic Graphs via Memory enhanced Normality}

\author{
\IEEEauthorblockN{Jie Liu}
\IEEEauthorblockA{School of Computer Science\\
    Northwestern Polytechnical University\\
    Xi'an, China\\
    jayliu@mail.nwpu.edu.cn}\\
\and
\IEEEauthorblockN{Xuequn Shang$^*$}
\IEEEauthorblockA{School of Computer Science\\
    Northwestern Polytechnical University\\
    Xi'an, China\\
    shang@nwpu.edu.cn}\\
\and
\IEEEauthorblockN{Xiaolin Han}
\IEEEauthorblockA{School of Computer Science\\
    Northwestern Polytechnical University\\
    Xi'an, China\\
    xiaolinh@nwpu.edu.cn}\\
\and
\IEEEauthorblockN{Kai Zheng}
\IEEEauthorblockA{School of Electrical Engineering \& Computer Science\\ 
    University of Electronic Science and Technology of China\\
    Chengdu, China\\
    zhengkai@uestc.edu.cn\\}
\and
\IEEEauthorblockN{Hongzhi Yin$^*$}
\IEEEauthorblockA{
School of Electrical Engineering \& Computer Science\\
The University of Queensland\\
    Brisbane, Australia\\
    h.yin1@uq.edu.au\\}
}

\maketitle

\begin{abstract}
Anomaly detection in dynamic graphs presents a significant challenge due to the temporal evolution of graph structures and attributes. The conventional approaches that tackle this problem typically employ an unsupervised learning framework, capturing normality patterns with exclusive normal data during training and identifying deviations as anomalies during testing. However, these methods face critical drawbacks: they either only depend on proxy tasks for representation without directly pinpointing normal patterns, or they neglect to differentiate between spatial and temporal normality patterns. More recent methods that use contrastive learning with negative sampling also face high computational costs, limiting their scalability to large graphs. To address these challenges, we introduce a novel Spatial-Temporal memories-enhanced graph autoencoder (\name). Initially, \name employs Graph Neural Networks (GNNs) and gated temporal convolution layers to extract spatial and temporal features. Then \name incorporates separate spatial and temporal memory networks to capture and store prototypes of normal patterns, respectively. These stored patterns are retrieved and integrated with encoded graph embeddings through a mutual attention mechanism. Finally, the integrated features are fed into the decoder to reconstruct the graph streams which serve as the proxy task for anomaly detection. This comprehensive approach not only minimizes reconstruction errors but also emphasizes the compactness and distinctiveness of the embeddings w.r.t. the nearest memory prototypes. Extensive experiments on six benchmark datasets demonstrate the effectiveness and efficiency of \name, where \name significantly outperforms existing methods with 5.8\% improvement in AUC scores and 4.62$\times$ faster in training time.
\end{abstract}


\footnotetext[1]{Corresponding authors.}

\section{Introduction}
% In recent years, graph anomaly detection has attracted increasing research attention due to its critical applications in domains such as social networks~\cite{ahmed2022combining}, traffic analysis~\cite{han2020traffic}, financial risk management~\cite{cheng2020graph}, and biological networks~\cite{mahmud2021deep}, to name a few. However, the majority of existing studies concentrate on static graphs, overlooking the fact that real-world graph data often evolve over time~\cite{zheng2021generative, ding2019deep}.

\begin{figure}[t]
    \centering
    \subfigure{
    \includegraphics[scale=0.33]{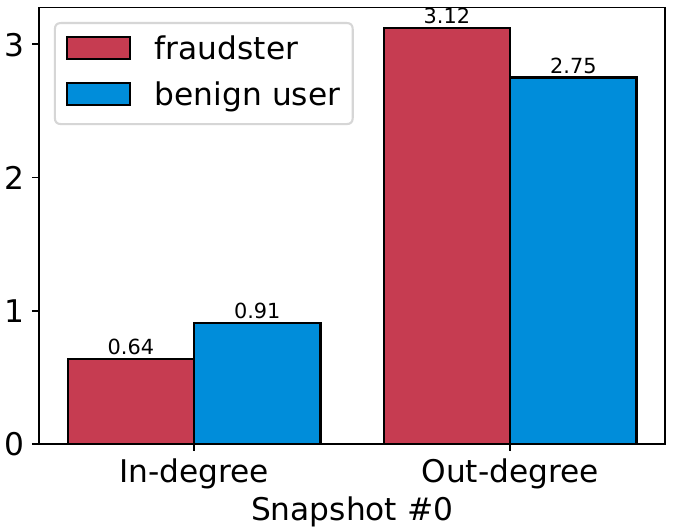}
    }\hspace{-1mm}
    \subfigure{
    \includegraphics[scale=0.33]{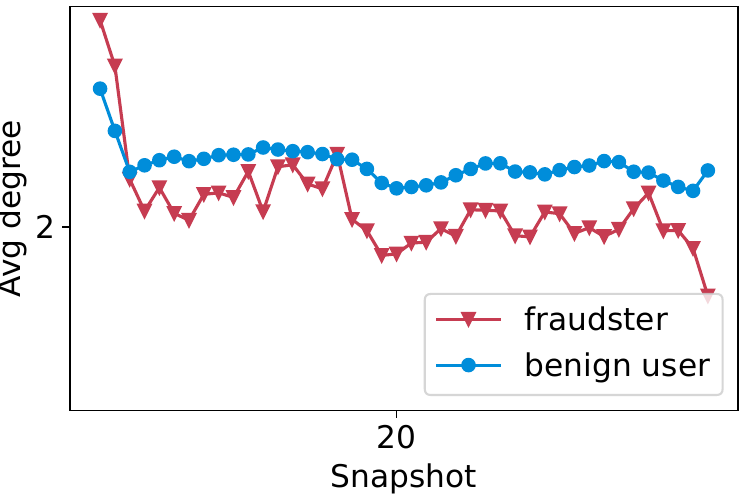}
    }
    \caption{Statistical observations of dynamic graphs on DGraph dataset. \textbf{Left}: Average degrees of fraudsters and benign users on snapshot \#0.
    \textbf{Right}: Degree curves of fraudsters and benign users as time evolves. Anomalous samples typically exhibit a higher frequency of vibrations, indicating that fraudsters frequently change their connections to other nodes.}
    \label{fig:toy}
\end{figure}

Real-world networks are often modeled as dynamic graphs to capture the changing nature of objects and their interactions~\cite{han2020traffic, zheng2019addgraph, zhang2021double, zheng2016keyword}. Beyond basic topology structure and node attributes, dynamic graphs encompass rich temporal signals, such as evolving patterns in graph structure and node attributes~\cite{ranshous2016scalable, yu2018netwalk, yang2023time}. This temporal dimension provides an additional perspective for analyzing anomalies. For instance, as illustrated in Fig. \ref{fig:toy}, anomalies may not be apparent when considering only the spatial information in a single snapshot, due to the similar degrees of fraudsters and normal users. However, observing temporal changes in graph structures can make abnormalities with higher vibration frequencies distinctly noticeable.

% For example, as shown in Fig. \ref{fig:toy}, if only consider the spatial information within snapshot \#40, the anomalies generally exhibit lower degrees than normal nodes. However, by observing the temporal changes in graph structures, the abnormality exhibit another typical pattern with higher vibration frequency.
To avoid ambiguity, in this paper, we define the structures and attributive features within individual graph snapshots as spatial information, while characterizing the evolving changes and trends among different snapshots as temporal information. To model and integrate both the spatial and temporal signals of nodes and edges for anomaly detection, there has been a growing interest in the study of anomaly detection in dynamic graphs~\cite{ma2021comprehensive, wang2021decoupling}.
% \begin{figure}[t!]
%     \centering
%     \includegraphics[scale=0.6]{toy.pdf}
%     \caption{A toy example of anomalies in dynamic graphs. The black nodes denote the normal nodes, and the red node denotes the abnormality, respectively. The anomaly in the graph becomes noticeable only when analyzing its temporal structural changes rather than just its spatial information from the $t$ snapshot.}
%     \label{fig:toy}
% \end{figure}

Due to the challenge of annotating anomalous objects in real-world scenarios, anomaly detection approaches for dynamic graphs mostly employ an unsupervised learning framework~\cite{liu2023bourne}. The key intuition behind these methods is to develop a model that captures patterns of normality by exclusively incorporating normal data during the training phase. 
Subsequently, objects that the model fails to accurately represent in the testing phase are classified as anomalies. 
% In the testing phase, models learned on normal data cannot compress abnormal data that have not been seen before, leading to large prediction or clustering errors.
For example, AddGraph~\cite{zheng2019addgraph} employs stacked GCN and GRU to capture spatial and temporal representations of normal edges and train an edge anomaly detector with link prediction as the proxy task. Then the edges with higher prediction errors in the test set are considered abnormal. TADDY~\cite{liu2021anomaly} 
uses a graph transformer to encode the representation of dynamic graphs in the training phase. The anomalous edges are detected based on a link prediction proxy task similar to \cite{zheng2019addgraph} in the test phase. NetWalk~\cite{yu2018netwalk} adopts a random-walk-based encoder to learn node representations and measures the node anomaly score by concerning its closest distance to the normal cluster centers. MTHL~\cite{teng2017anomaly} projects multi-view dynamic graphs into a shared latent subspace and learns a compact hypersphere surrounding normal samples. Node anomalies are detected based on the distance to the learned hypersphere center. OCAN~\cite{zheng2019one} captures the normal activity patterns of observed benign users’ attributes and detects fraudsters that behave significantly differently. 

Despite their success, these methods face several limitations: (1) Methods like AddGraph\cite{zheng2019addgraph}, TADDY~\cite{liu2021anomaly}, StrGNN\cite{cai2021structural} only leverage proxy tasks (e.g., edge stream prediction) to derive feature representation, rather than directly identifying normal patterns. This can be ineffective if anomalies share structural or attributive similarities with normal data, due to the powerful representation learning models (e.g., GCN, Transformer) used. Consequently, abnormal data might also be well-represented, resulting in suboptimal anomaly detection (\textbf{P1}). (2) Approaches like NetWalk~\cite{yu2018netwalk}, MTHL~\cite{teng2017anomaly}, SAD~\cite{tian2023sad} explicitly model normal patterns through clustering or hypersphere centers but do not distinguish between spatial and temporal patterns. However, as illustrated in Fig. \ref{fig:toy}, the prototypical pattern of nodes within a single snapshot can be different from the pattern of evolving trends of nodes among different snapshots, necessitating separate identification and storage of spatial and temporal normal patterns (\textbf{P2}). (3) State-of-the-art methods such as SAD~\cite{tian2023sad} and CLDG~\cite{xu2023cldg} employ contrastive learning to build self-supervised frameworks, relying on extensive negative sampling to prevent model collapsing. However, the generation and processing of a substantial number of negative samples in large-scale graphs will bring high computational costs and reduce training efficiency  (\textbf{P3}).

In order to address all the limitations above, we propose a novel \underline{\textbf{S}}patial-\underline{\textbf{T}}emporal memo\underline{\textbf{RI}}es enhanced gra\underline{\textbf{P}}h auto\underline{\textbf{E}}ncoder framework (\textbf{\name} for abbreviation) for node anomaly detection in dynamic graphs. The key idea behind \name is to leverage two separate memory networks~\cite{weston2014memory} to identify and preserve the spatial and temporal patterns of normality and integrate them with a graph autoencoder to reconstruct graph streams as the proxy task for anomaly detection. Specifically, spatial and temporal node embeddings from input graph streams are derived using Graph Neural Networks (GNNs) and gated temporal convolution, serving as the spatial and temporal encoders, respectively. The spatial and temporal patterns are then written into their respective memory banks via a mutual attention mechanism on node embeddings, with each memory item encapsulating a prototype of normal node patterns. After that, the encoded spatial and temporal embeddings access the most closely related prototypes within the memory through mutual attention-driven retrieval. These retrieved items are subsequently merged and combined with the initial embeddings and fed into the decoder to reconstruct the graph streams.

During training, spatial and temporal memory items are updated along with the encoder and decoder. We propose a comprehensive training objective that minimizes both reconstruction errors and compactness errors, promoting proximity between node embeddings and their nearest memory items. Additionally, we also minimize separateness errors to enhance the distinctiveness of memory items. This ensures \name's effective use of a limited number of memory items, significantly reducing the size of the memory bank and enhancing model efficiency. Moreover, since \name is non-contrastive learning model, it requires no negative sampling or data augmentation, which also helps promote model scalability.

In the testing phase, the learned memory items remain fixed, and the reconstruction and compactness loss now serve as the anomaly score. Since the reconstruction process integrates normality patterns preserved in memory, the inputs that deviate from these prototypical patterns of normal data are likely to yield elevated anomaly scores, thereby facilitating their identification as anomalies.

In summary, the main contributions of our work are as follows:
\begin{itemize}
    \item We propose a novel spatial-temporal memory-enhanced graph autoencoder framework, \name, that explicitly captures normality patterns and integrates them into graph stream reconstruction for anomaly detection. 
    % To the best of our knowledge, this is the first work to integrate memory networks for anomaly detection in dynamic graphs.
    \item Considering the distinct normality patterns in spatial and temporal dimensions, we develop two independent memory modules that can capture and preserve spatial and temporal patterns separately. To measure the complex relations between node embeddings and diverse spatial and temporal memory items, we propose a mutual attention mechanism to update and retrieve memory items.
    \item We propose an efficient dynamic anomaly detection model by promoting separateness among
    memory items, thereby significantly reducing the size of memory bank.
    \item Extensive experiments on six benchmark datasets have demonstrated the state-of-the-art performance of \name, achieving an average AUC score improvement of 5.8\%. Both theoretical analysis and empirical results highlight \name's efficiency, showing linear scalability with the increase of node numbers.
\end{itemize}

\section{Related Work}
In this section, we introduce the works closely related to ours: Anomaly Detection in static graphs, Anomaly Detection in Dynamic Graphs and Memory Networks.
\subsection{Anomaly Detection in static graphs}
Graph anomaly detection aims at identifying anomalous graph objects (i.e., nodes, edges, and subgraphs) in the graph~\cite{ma2021comprehensive, wang2021decoupling, huang2023unsupervised}. 
Early anomalous node detection approaches mainly use shallow techniques such as residual analysis (Radar~\cite{li2017radar}), matrix factorization (ALAD~\cite{liu2017accelerated}), and CUR decomposition (ANOMALOUS~\cite{peng2018anomalous}) to extract anomalous pattern in graphs. More recently, DOMINANT~\cite{ding2019deep} pioneered the integration of deep learning into node anomaly detection by employing a graph autoencoder~\cite{he2024ada} to reconstruct both the structure and attribute information of graphs. CoLA~\cite{liu2021anomaly} SL-GAD~\cite{zheng2021generative} and CONAD~\cite{xu2022contrastive} further introduce graph contrastive learning that captures abnormal patterns by measuring agreement between augmented item pairs.
\subsection{Anomaly Detection in Dynamic Graphs}
Recently, the field of anomaly detection in dynamic graphs has garnered significant attention, primarily because of its capability to identify abnormalities in graphs that exhibit time-varying characteristics.

Within dynamic networks, the definition of an anomalous object varies widely depending on the specific application context. Based on the diverse nature of anomalies that can occur in such evolving structures, the scope of detection tasks can range from identifying abnormal nodes~\cite{teng2017anomaly, ji2013incremental, liu2023meta, liu2023imbalanced} and edges~\cite{aggarwal2011outlier, sricharan2014localizing, ranshous2016scalable, manzoor2016fast} to pinpointing anomalous subgraphs~\cite{chen2012community, eswaran2018spotlight}. Early approaches mainly leverage the shallow mechanisms to detect anomalies in dynamic graphs. For example, 
% CAD~\cite{sricharan2014localizing} localizes anomalous edges by tracking a measure that combines information regarding changes in graph structure and changes in edge weights. 
CM-sketch~\cite{ranshous2016scalable} utilizes sketch-based approximation of structural properties of the graph stream to calculate edge outlier scores. MTHL~\cite{teng2017anomaly} distinguishes normal and anomalous nodes according to their distances to the learned hypersphere center. SpotLight~\cite{eswaran2018spotlight}  guarantees a large mapped distance between anomalous and normal graphs in the sketch space with a randomized sketching technique. 

More recently, another branch of methods employs deep learning techniques to capture anomalous objects in dynamic graphs~\cite{zheng2019addgraph,yu2018netwalk, cai2021structural,liu2021anomaly}. NetWalk~\cite{yu2018netwalk} utilizes a random walk-based encoder to generate node embeddings and score the abnormality of nodes and edges with their distance to cluster centers. AddGraph~\cite{zheng2019addgraph} employs stacked GCN and GRU to capture spatial and temporal representations of normal edges and train an edge anomaly detector with edge stream prediction as the proxy task. StrGNN~\cite{cai2021structural} further extracts the h-hop enclosing subgraph for each edge and employs stacked GCN and GRU to encode the extracted subgraphs for edge stream prediction.  TADDY~\cite{liu2021anomaly2} learns the representations from dynamic graphs with coupled spatial-temporal patterns via a transformer. SAD~\cite{tian2023sad} and CLDG~\cite{xu2023cldg} introduce contrastive learning for anomaly detection in dynamic graphs.

Most of the above approaches either only depend on proxy tasks for general representation without directly pinpointing normal patterns, or they neglect to differentiate between spatial and temporal normality patterns, leading to diminished efficacy in anomaly detection. \name alleviates this problem by capturing distinct spatial and temporal normality patterns in the training phase and integrating the preserved normality patterns to detect anomalies in the test phase.

\subsection{Memory Networks}
To address the challenge of capturing long-term dependencies in temporal data, researchers recently proposed memory networks~\cite{weston2014memory}. These networks can read and write to global memories where individual items in the memory correspond to prototypical patterns of the features. MemN2N~\cite{sukhbaatar2015end} further enhances memory networks to operate in an end-to-end manner, which reduces the need for layer-wise supervision during training. Memory networks have shown effectiveness in various memorization tasks ranging from unsupervised feature learning~\cite{kim2018memorization, wu2018unsupervised}, one-shot learning~\cite{santoro2016meta, cai2018memory}, to image generation~\cite{zhu2019dm}. Recognizing the memory's ability to capture and store prototypical patterns of normal data, more recent studies have started combining Autoencoders~\cite{bengio2006greedy,kipf2016variational} with memory modules to detect anomalies in video~\cite{gong2019memorizing, park2020learning} and graph~\cite{niu2023graph} data.

However, the focus of these methods has largely been on video or static graph data. Our work differs by applying memory networks to dynamic graphs. We have developed distinct spatial and temporal memory modules, which allow us to analyze normal prototypes in both spatial and temporal dimensions independently.

\section{Preliminaries}
In this section, we provide the definitions of essential concepts and formalize the problem of dynamic graph anomaly detection. 
% We also summarize the frequently used notations across this paper in Table \ref{tab:nota} for quick reference. 

\xhdr{Definition 1: Dynamic Graph} 
Given a dynamic graph with overall timestamps of T, we use $\mathbb{G}=\{\mathcal{G}^{t}\}^T_{t=1}$ to denote the graph stream, where each $\mathcal{G}^t=\{\mathcal{V}^t, \mathcal{E}^t\}$ is the snapshot at timestamp $t$. $\mathcal{V}^t$ and $\mathcal{E}^t$ is the node set and edge set at timestamp $t$. An edge $e^t_{i,j}=(v^t_i, v^t_j)\in\mathcal{E}^t$ indicates the connection between node $v^t_i$ and $v^t_j$ at timestamp $t$, where $v^t_i, v^t_j\in \mathcal{V}^t$. $N_t=|\mathcal{V}^t|$ and $M_t=|\mathcal{E}^t|$ indicate the number of nodes and edges in timestamp $t$. The structural information of $\mathcal{G}^t$ is represented by the graph adjacency matrix $\mathbf{A}^t\in\mathbb{R}^{N_t\times N_t}$, where $\mathbf{A}^t_{ij}=1$ if $e^t_{ij}$ exists, otherwise $\mathbf{A}^t_{ij}=0$. $\mathbf{X}^t\in\mathbb{R}^{N_t\times D}$ denotes the node feature matrix at timestamp $t$ and its $i$-th row vector $\mathbf{x}^t_i\in\mathbb{R}^D$ represents the feature of node $v^t_i$.

\xhdr{Definition 2: Anomaly Detection in Dynamic Graphs}
Given a dynamic graph $\mathbb{G}=\{\mathcal{G}^t\}^T_{t=1}$, our goal is to lean an anomaly detection function $f(\cdot):\mathbb{R}^{N_t\times D}\rightarrow\mathbb{R}^{N_t\times 1}$ that can measure the degree of abnormality of the node by calculating its anomaly score $f(v^t_{i})$. A larger anomaly score indicates a higher abnormal probability for $v^t_{i}$.

Given the challenges in obtaining and accessing anomalous labels in real-world networks, we adopt an unsupervised setting for detecting anomalies in dynamic graphs. In the training phase of this research, no node labels that indicate abnormalities are used. Instead, it is presumed that all nodes present during training exhibit normal behavior. The binary labels denoting abnormality are introduced only in the testing phase to assess the model's effectiveness. In this context, a label $y_{v^t_{i}}=0$ signifies that the node $v^t_{i}$ is considered normal, whereas $y_{v^t_{i}}=1$ identifies it as abnormal.

\section{Methodology}
\begin{figure*}[t!]
    \centering
    \includegraphics[scale=0.62]{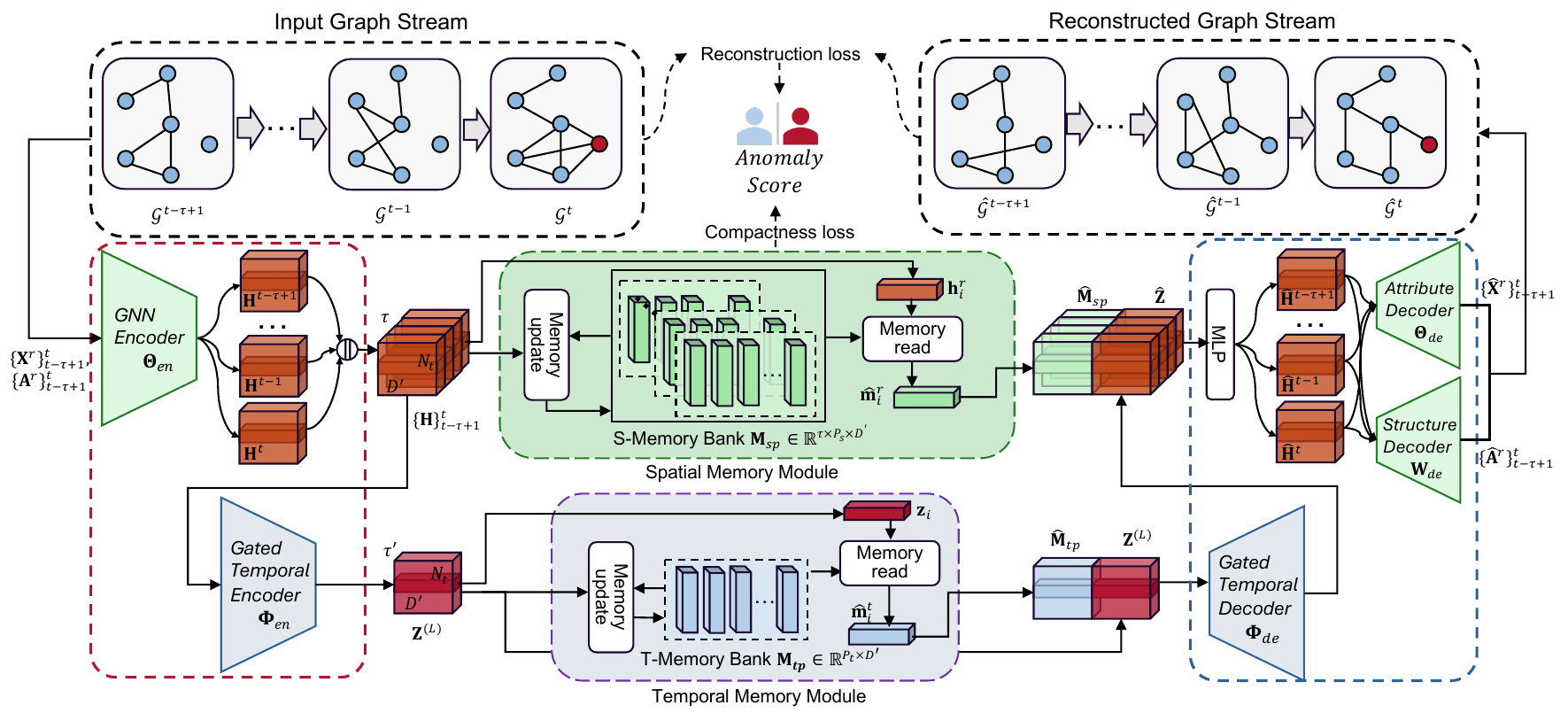}
    \caption{Overall framework of the proposed \name.}
    \label{fig:frame}
\end{figure*}

In this section, we present our proposed framework, \name, designed for node anomaly detection within dynamic graphs in an unsupervised manner. As depicted in Fig. \ref{fig:frame}, \name is comprised of four key components: (1) \textit{Spatial-temporal graph encoder} that encodes both the spatial and temporal information of input graph stream into comprehensive node embeddings. (2) \textit{Spatial-temporal memory learning} that captures and stores the prototypical patterns of normal node representations at both spatial and temporal dimensions. (3) \textit{Spatial-temporal graph decoder} that reconstructs the original graph stream using the latent node embeddings and the identified normal prototypes, facilitating the comparison with the original input. (4) \textit{Unified anomaly detector} that measures the abnormality of a node by calculating the reconstruction errors between the original and reconstructed graphs and the compactness errors between the node and its nearest prototype.

In the rest of this section, we introduce the four components in detail from section \ref{sub:auto} to \ref{sub:detec}. The overall pipeline of \name is illustrated in Algorithm (\ref{alg:1}).

\subsection{Spatial-Temporal Graph Encoder}\label{sub:auto}
The input dynamic graph contains not only structural and attributive information within each graph snapshot but also abundant temporal information illustrating the evolution alongside the graph stream. Capturing both the spatial and temporal properties of dynamic graphs is essential for detecting anomalies. To address this challenge, we design a spatial-temporal encoder that consists of a spatial encoder and a temporal encoder. 

\subsubsection{Spatial Encoder}
GNNs have recently emerged as one of the most powerful network representation learning approaches due to their ability to conduct deep learning on non-Euclidean graph data. In this work, we employ an L-layer GNN as the spatial encoder. Instead of inputting the whole graph stream at a time, we consider a graph sequence $\{\mathcal{G}^{t-\tau+1}, \ldots, \mathcal{G}^t\}$ over a time window size $\tau$. By adjusting the hyper-parameter $\tau$, we can specify the receptive fields along the time axis.

The spatial encoder takes the graph sequence $\{\mathcal{G}^{t-\tau+1}, \ldots, \mathcal{G}^t\}$ as input and outputs the latent node embeddings $\mathbf{H}^t\in\mathbb{R}^{N_t\times D'}$ for each snapshot $\mathcal{G}^t$. Specifically, node embeddings in $\mathcal{G}^t$ are computed as follows:
\begin{equation}\label{eq:gnn}
\mathbf{H}^{(t, l)}=\textit{GNN}\Bigl(\mathbf{A}^t, \mathbf{H}^{(t, l-1)}; \mathbf{\Theta}_{en}^{(l)}\Bigr),
\end{equation}
where $\mathbf{\Theta}_{en}^{(l)}\in\mathbb{R}^{D\times D'}$ denotes the learnable weight parameters of the $l$-th layer GNN. $\mathbf{H}^{(t, l-1)}$ and $\mathbf{H}^{(t, l)}$ are the node representation matrices learned by the $(l-1)$-th and $(l)$-th layer, respectively. $\mathbf{H}^{(t,0)}$ is $\mathbf{X}^t$. $\textit{GNN}_{\theta}(\cdot)$ can be set as any off-the-shelf graph neural networks. For computation efficiency, we adopt a two-layer graph convolutional network (GCN) as the backbone. Thus, Equation (\ref{eq:gnn}) can be specifically re-written as:
\begin{equation}\label{eq:gcn}
    \mathbf{H}^{(t, l)} = ReLU\Bigl(\widetilde{\mathbf{D}}^{-\frac{1}{2}}_t\widetilde{\mathbf{A}}^t\widetilde{\mathbf{D}}^{-\frac{1}{2}}_t\mathbf{H}^{(t, l-1)}\mathbf{\Theta}_{en}^{(l)}\Bigr),
\end{equation}
where $\widetilde{\mathbf{A}}^t=\mathbf{A}^t+\mathbf{I}_{N^t}$, and $\widetilde{\mathbf{D}}_t$ is a diagonal node degree matrix where $\widetilde{\mathbf{D}}_t(i,i)=\sum_j\widetilde{\mathbf{A}}_t(i,j)$. $\textit{ReLU}(\cdot)$ is the activation function. We simplify $\mathbf{H}^{(t,L)}$ as $\mathbf{H}^t$.

\subsubsection{Gated Temporal Encoder}
Having calculated the node embeddings for each graph snapshot, we next incorporate the temporal dependencies observed across different snapshots. Prior research~\cite{zheng2019addgraph, cai2021structural} predominantly employed Recurrent Neural Networks (RNNs) for learning temporal information. The inherent limitation of RNNs, however, is their sequential processing requirement for each time step, which significantly increases computational costs and reduces the efficiency of the model. To overcome this challenge, we utilize gated temporal convolution for temporal learning, which facilitates the parallel processing of elements within the graph sequence.

For gated temporal convolution, we employ a 1-dimensional convolution with a kernel width of $K_t$ to capture dynamic evolving between timestamps $t-\tau+1$ and $t$ of the graph sequence $\{\mathcal{G}^{t-\tau+1}, \ldots, \mathcal{G}^t\}$. Since the node embeddings from each snapshot influence anomaly detection differently~\cite{cai2021structural}, we incorporate a Gated Linear Unit (GLU) subsequent to the 1-dimensional convolution layer, which serves to accentuate critical information more significantly associated with anomaly detection.

Specifically, given $\mathbf{Z}^{(0)}=\{\mathbf{H}\}^t_{t-\tau+1}\in\mathbb{R}^{\tau\times N_t\times D'}$ as input, the gated temporal convolution is defined as follows:
\begin{align}\label{eq:gated}
    \mathbf{Z}^{(l)}&=tanh(\mathbf{E}_1)\odot\sigma(\mathbf{E}_2),\\ 
    \mathbf{E}_1&=\mathbf{E}^{(l-1)}[:, 1:D'],\\
    \mathbf{E}_2&=\mathbf{E}^{(l-1)}[:, D'+1:2D'],\\
    \mathbf{E}^{(l-1)}&=\mathbf{Z}^{(l-1)}\mathbf{\Phi}_{en}^{(l)}.
\end{align}
Where $\mathbf{\Phi}_{en}^{(l)}\in\mathbb{R}^{D'\times2D'}$ is the weight parameter of the 1-dimensional convolution kernel, $\odot$ denotes the element-wise multiplication. $\sigma(\mathbf{E})=\frac{1}{1+e^{-\mathbf{E}}}$ denotes the logistic sigmoid. After stacking L-layer of gated temporal convolution, the length of the graph sequence is reduced to $\tau'=\tau - L\times(K_t-1)$. We simplify $\mathbf{Z}^{(L)}\in\mathbb{R}^{\tau'\times N_t\times D'}$ as $\mathbf{Z}$.

\subsection{Spatial-Temporal Memory Learning}\label{sub:mem}
The spatial-temporal memory learning aims to capture and store prototypical spatial patterns and temporal patterns of normal node embeddings. Spatial and temporal memory banks contain $\tau P_s$ and $P_t$ memory items of dimension $D'$, denoted as $\mathbf{M}_{sp}\in\mathbb{R}^{\tau\times P_s\times D'}$ and $\mathbf{M}_{tp}\in\mathbb{R}^{P_t\times D'}$, respectively. 
% $\mathbf{M}_{sp}$ and $\mathbf{M}_{tp}$ share the same memory update and read procedures. For simplicity, we only present the learning process for $\mathbf{M}_{tp}$.
In the rest of this section, we introduce the spatial and temporal memory modules, respectively.
\subsubsection{Spatial Memory Module}
We denote $\mathbf{m}^{r}_p\in \mathbb{R}^{D'} (p=1,\ldots,P_s;r=t-\tau+1,\ldots,t)$ as the item of memory $\mathbf{M}_{sp}$, and $\mathbf{h}^r_i\in\mathbb{R}^{D'} (i=1, \dots, N_t; r=t-\tau+1,\ldots,t)$ as the spatial encoded feature of node $i$ at time $r$.

\begin{figure}[t!]
  \centering
  \subfigure[memory read.]{
  \includegraphics[scale=0.5]{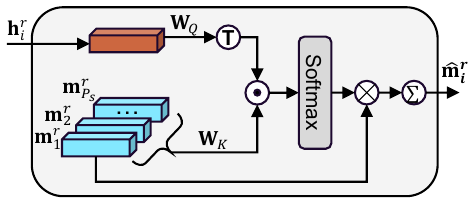}
  \label{fig:read}}
  \subfigure[memory update.]{
  \includegraphics[scale=0.5]{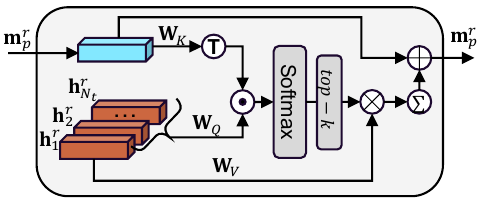}
  \label{fig:update}}
  \caption{The illustration of (a) memory read and (b) memory update procedures in the spatial memory module.} 
\end{figure}

\xhdr{Memory Read.}
As shown in Fig. \ref{fig:read}, the reading process begins by calculating the attention weights between each node feature $\mathbf{h}^r_i$ and all memory items $\mathbf{m}^r_p$. Prior research~\cite{niu2023graph,liu2022learning} primarily adopts cosine similarity to compute self-attention, which restricts the capability to explore the relations between node features and diverse spatial and temporal memory items. To address this problem, we employ a mutual attention mechanism:
\begin{align}
    \mathbf{k}_p&=\mathbf{m}^r_p\mathbf{W}_K,\\
    \mathbf{q}_i&=\mathbf{h}^r_i\mathbf{W}_Q,\\
    \mathbf{v}_i&=\mathbf{h}^r_i\mathbf{W}_V,
\end{align}
where $\mathbf{W}_K$, $\mathbf{W}_Q$ and $\mathbf{W}_V\in\mathbb{R}^{D'\times D'}$ are weight matrices of key vector, query vector, and value vector, respectively. The attention weights $w_{(i,p)}$ are then computed with softmax function:
\begin{equation}\label{eq:w}
    w_{(i, p)} = \frac{\text{exp}\bigl(\mathbf{q}_i(\mathbf{k}_p)^T\cdot\frac{1}{\sqrt{D'}}\bigr)}{\sum^{P_s}_{p'=1}\text{exp}\bigl(\mathbf{q}_i(\mathbf{k}_{p'})^T\cdot\frac{1}{\sqrt{D'}}\bigr)}.\\
\end{equation}
For each node feature $\mathbf{h}^r_i$, we read the memory by a weighted average of the items $\mathbf{m}^r_p$ with the corresponding weights $w_{(i, p)}$, and obtain the readout memory item $\hat{\mathbf{m}}^r_i\in \mathbb{R}^{D'}$ as follows:
\begin{equation}\label{eq:mi}
    \hat{\mathbf{m}}^r_i=\sum_{p=1}^{P_s} w_{(i, p)}\mathbf{m}^r_p,
\end{equation}
Assigning this procedure to all $i\in [1,N_t]$ and $r\in[t-\tau+1,t]$, we obtain the readout memory matrix $\hat{\mathbf{M}}_{sp}\in\mathbb{R}^{\tau\times N_t\times D'}$.

\xhdr{Memory Update}
During the training phase, the memory bank will also be updated to record the spatial prototypes of normal nodes. As shown in Fig. \ref{fig:update}, for each memory item $\mathbf{m}^r_p$, we select the node features that are nearest to it based on the matching weights $\mu_{(i, p)}$ as follows:
\begin{equation}\label{eq:mu}
    \mu_{(i, p)} = \frac{\text{exp}\bigl(\mathbf{q}_i(\mathbf{k}_p)^T\cdot\frac{1}{\sqrt{D'}}\bigr)}{\sum^{\tau'N_t}_{i'=1}\text{exp}\bigl(\mathbf{q}_{i'}(\mathbf{k}_{p})^T\cdot\frac{1}{\sqrt{D'}}\bigr)}.\\
\end{equation}  
Contrary to \cite{niu2023graph, park2020learning}, which utilizes all features for updating memory items, we selectively employ only the top-$K$ relevant features. This strategy effectively filters out irrelevant, noisy nodes, thereby capturing and recording the general patterns of normal events more effectively. Therefore, the top-$K$ values of $\mu_{(i, p)}$ are preserved, while the remainder is nullified to 0. The updated memory item $\mathbf{m}_p$ is then calculated as follows:
\begin{equation}\label{eq:mp}
    \mathbf{m}^r_p\leftarrow \mathbf{m}^r_p + \sum_{i=1}^{K}\mu_{(i, p)}\mathbf{v}_i.
\end{equation}

\subsubsection{Temporal Memory Module}
Unlike spatial memories that capture normal patterns within each graph snapshot, temporal memories aim to characterize the prototypical pattern of evolving trends among different snapshots.
Specifically, $m_q\in\mathbb{R}^{D'}$ is the item of memory bank $\mathbf{M}_{tp}$. For each temporal node feature $\mathbf{z}_i\in\mathbb{R}^{D'} (i=1, \dots, \tau'\times N_t)$, its corresponding readout memory $\hat{\mathbf{m}}_i$ is calculated as follows:
\begin{equation}\label{eq:mi2}
    \hat{\mathbf{m}}_i=\sum_{q=1}^{P_t} w_{(i, q)}\mathbf{m}_q.
\end{equation}
We adopt the same memory read procedure as in Eq.~(\ref{eq:w}) to calculate $w_{(i, q)}$ and utilize Eq.~(\ref{eq:mu}) to update temporal memories. The overall readout matrix for temporal features is denoted as $\hat{\mathbf{M}}_{sp}\in\mathbb{R}^{\tau\times N_t\times D'}$. Each item within these matrices represents the averaged spatial and temporal normal prototypes associated with the corresponding node. 

\subsection{Spatial-temporal Graph Decoder}
% After the memory read procedure in both spatial and temporal memory modules, we obtain the averaged memory matrices $\hat{\mathbf{M}}_{tp}\in\mathbb{R}^{\tau'\times N_t\times D'}$ and $\hat{\mathbf{M}}_{sp}\in\mathbb{R}^{\tau\times N_t\times D'}$. Each item within these matrices represents the averaged spatial and temporal normal prototypes associated with the corresponding node. 
In this section, we reconstruct the original graph stream with the averaged memory matrices and latent representation $\mathbf{Z}$. As shown in Fig. \ref{fig:frame}, our method inputs the concatenation of $\hat{\mathbf{M}}_{tp}$ and $\mathbf{Z}$ into the gated temporal decoder, which is the reverse process of equation (\ref{eq:gated}), and outputs $\hat{\mathbf{Z}}\in\mathbb{R}^{\tau\times N_t\times D'}$. Then we concatenate $\hat{\mathbf{M}}_{sp}$ with $\hat{\mathbf{Z}}$ and input it into a one-layer MLP to obtain $\{\hat{\mathbf{H}}\}^t_{t-\tau+1}$:
\begin{equation}
    \hat{\mathbf{H}}^t=MLP\bigl(\hat{\mathbf{M}}_{sp}[t,:,:]\parallel\hat{\mathbf{Z}}[t, :, :]\bigr), \\
\end{equation}
where $\parallel$ denotes concatenation operation and $\hat{\mathbf{H}}^t\in\mathbb{R}^{N_t\times D'}$. Then, we use an L-layer GCN as the attributive decoder to reconstruct attribute matrix $\hat{\mathbf{X}}^{t}$:
\begin{equation}\label{eq:attr_dec}
\hat{\mathbf{X}}^{t}=\textit{GCN}\Bigl(\mathbf{A}^t, \hat{\mathbf{H}}^t; \mathbf{\Theta}_{de}^{(l)}\Bigr),
\end{equation}
We employ the inner product of $\hat{\mathbf{H}}^t$ as the structural decoder, formatted as follows:
\begin{equation}\label{eq:stru_dec}
\hat{\mathbf{A}}^t=\sigma\Bigl(\hat{\mathbf{H}}^t\mathbf{W}_{de}\bigl(\hat{\mathbf{H}}^t\bigr)^T\Bigr).
\end{equation}
Where $\mathbf{W}_{de}\in\mathbb{R}^{D'\times D'}$ is the weight matrix and $\sigma(\cdot)$ denotes the Sigmoid activation function.

\subsection{Unified Anomaly Detector}\label{detec}
In the previous sections, we have calculated the spatial and temporal prototypes and integrated them with latent representations to reconstruct the original graph stream. For a training timestamp $t$, the reconstruction errors can be formatted as a combination of attributive and structural reconstruction errors:
\begin{equation}\label{eq:rec_loss}
    \mathcal{L}_{rec}=\alpha\Vert\hat{\mathbf{X}}^t-\mathbf{X}^t\Vert_2+(1-\alpha)\Vert\hat{\mathbf{A}}^t-\mathbf{A}^t\Vert_2,
\end{equation}
where $\alpha\in[0,1]$ is a hyper-parameter that balances the importance of attributive and structural errors.

Given that only normal nodes are present during the training phase, under ideal circumstances, the features of a normal node should be close to the nearest item in the memory. Conversely, the features of abnormal nodes are expected to be distant from any memory items, reflecting their deviation from normal patterns. Encouraged by this, the feature compactness loss, denoted by $\mathcal{L}_{com}$, is as follows:
\begin{equation}\label{eq:com_loss}
    \mathcal{L}_{com}=\sum^{\tau N_t}_{i=1}\Vert\mathbf{h}_i-\mathbf{m}^{sp}_p\Vert_2+\sum^{\tau' N_t}_{i=1}\Vert\mathbf{z}_i-\mathbf{m}^{tp}_p\Vert_2,
\end{equation}
where $\mathbf{m}^{sp}_p$ and $\mathbf{m}^{tp}_p$ denote the spatial and temporal memory item that is nearest to spatial embedding $\mathbf{h}_i$ and temporal embedding $\mathbf{z}_i$, respectively.

Furthermore, the items within the memory should be sufficiently distant from each other. This spacing ensures that a broad spectrum of normal data patterns can be effectively captured and represented. Encouraged by this, we design the memory separateness loss, denoted as $\mathcal{L}_{sep}$, as follows:
\begin{align}\label{eq:sep_loss}
\begin{split}
    \mathcal{L}_{sep}&=\sum^{\tau N_t}_{i=1}\bigl[\Vert\mathbf{h}_i-\mathbf{m}^{sp}_p\Vert_2-\Vert\mathbf{h}_i-\mathbf{m}^{sp}_n\Vert_2\bigr]_+\\ 
    &+ \sum^{\tau' N_t}_{i=1}\bigl[\Vert\mathbf{z}_i-\mathbf{m}^{tp}_p\Vert_2 - \Vert\mathbf{z}_i-\mathbf{m}^{tp}_n\Vert_2\bigr]_+,
\end{split}
\end{align}
where $\mathbf{m}^{sp}_p$ and $\mathbf{m}^{sp}_n$ denotes the nearest and the second nearest memory item to $\mathbf{h}_i$. $\mathbf{m}^{tp}_p$ and $\mathbf{m}^{tp}_n$ denote the nearest and the second nearest memory item to $\mathbf{z}_i$.
The total loss for training is formatted as:
\begin{equation}\label{eq:loss}
\mathcal{L}=\mathcal{L}_{rec}+\mathcal{L}_{com}+\mathcal{L}_{sep}.
\end{equation}
During inference, the sum of $\mathcal{L}_{rec}$ and $\mathcal{L}_{com}$ is adopted as the anomaly score. The anomaly score for each node is calculated
$R$ times to ensure that the final anomaly scores are statistically stable and reliable.

\begin{algorithm} [t!]
  \caption{Forward propagation of \name }\label{alg:1}
  \KwIn{Graph stream $\{\mathcal{G}^t\}^T_{t=1}$; Number of training epochs $I$; Time window size $\tau$; Temporal convolution kernel width $K_t$; Evaluation rounds $R$.\\}
  \KwOut{Anomaly scoring function $f(\cdot)$.\\} 
  Randomly initialize the trainable parameters of the encoder, decoder, memory modules, and scoring function;\\
  \tcc{Training stage}
  \For{$i\in1,2,\ldots,I$}{
  \For{snapshot $\mathcal{G}^t\in\{\mathcal{G}^t\}^T_{t=1}$}{
  Extract the graph sequence $\{\mathcal{G}^{t-\tau+1}, \ldots, \mathcal{G}^t\}$; \\
  Calculate spatial node embeddings $\mathbf{H}^t$ via Eq. (\ref{eq:gnn});\\
  Calculate temporal node embeddings $\mathbf{Z}$ via Eq. (\ref{eq:gated});\\
  \For{$p\in1,2,\ldots,P$}{
  Calculate the read attention weights $w_{(i,p)}$ via Eq. (\ref{eq:w}) and update attention weights $\mu_{(i,p)}$ via Eq. (\ref{eq:mu});\\
  }
  Readout the averaged memories via Eq. (\ref{eq:mi2});\\
  Update the memory items via Eq. (\ref{eq:mp});\\
  Calculate the reconstructed attributes $\hat{\mathbf{X}}^t$ and structures $\hat{\mathbf{A}}^t$ via Eq. (\ref{eq:attr_dec}) and Eq. (\ref{eq:stru_dec});\\
  Compute the loss objective via Eq. (\ref{eq:rec_loss}), (\ref{eq:com_loss}), and (\ref{eq:sep_loss}).\\
  }
  }
  \tcc{Inference Stage}
  \For{$r\in1,2,\ldots,R$}{
  \For{$v_i\in\{\mathcal{V}^t\}^T_{t=1}$}{
  Calculate the anomaly score for each node $v_i$ via $\mathcal{L}_{rec}$ and $\mathcal{L}_{com}$.\\
  }
  }
\end{algorithm}

\subsection{Complexity Analysis}\label{sec:complix}
In this subsection, we analyze the time complexity of each component in the \name framework. We employ an L-layer GCN for spatial encoding and decoding of a graph sequence with window size $\tau$, which brings a complexity of $\mathcal{O}(\tau LM_tD'+\tau LN_tD'^2)$, where $M_t$ and $N_t$ are the averaged edge and node number for each snapshot. For memory read and update, the complexity is mainly caused by the mutual attention mechanism between node features and memory items, which is $\mathcal{O}(\tau N_tP_tD')$. The time complexity for an L-layer gated temporal convolution is $\mathcal{O}(\tau LN_tD'^2)$. Therefore, to apply an L-layer \name to a graph sequence of window size $\tau$, the overall time complexity is $\mathcal{O}\bigl(\tau L(M_tD'+N_tD'^2)+\tau N_tP_tD')\bigr)$. As investigated in section \ref{subsub:n_item}, $P_t$ can be restricted to a very limited number, thereby the time complexity of \name is approximately linear to node number. Section \ref{sub:eff} provide an empirical analysis of model efficiency.

\section{Experimental Study}
In this section, we conduct extensive experiments on six real-world benchmark datasets to evaluate the performance of \name. Specifically, from section\ref{sub:dataset} to \ref{sub:para}, we introduce the experimental setups. Then in section \ref{sub:detec} and \ref{sub:rate}, we compare our method with the state-of-the-art baseline methods on node anomaly detection task. We then evaluate time efficiency of the model in section \ref{sub:eff} and conduct ablation study to validate the effectiveness of each component of \name in section \ref{sub:ablation}. In section \ref{sub:sens}, we study the parameter sensitivity to further investigate the property of \name. We also demonstrate the effectiveness of the proposed spatial and temporal memory modules with a case study in \ref{sub:case}.

\subsection{Datasets}\label{sub:dataset}
We assess the performance of \name and its competitors on six real-world temporal networks. The description of the datasets is shown in Table~\ref{tab:datasets}. Among them, DBLP-3 and DBLP-5\footnote{\url{https://dblp.uni-trier.de}} are co-author networks that consist of authors as nodes and co-authorship as edges, with the node features being abstracts of the author’s publications during certain period encoded by word2vec. The authors in DBLP-3 and DBLP-5 are from three and five research areas, respectively. Reddit\footnote{\url{https://www.reddit.com/}} is a social network where the nodes represent the posts and edges are defined through similar keywords. Word2vec is applied to the comments of a post to generate its node attributes. Brain is a biological network, with nodes symbolizing distinct cubes of brain tissue and edges reflecting their connectivity. 

\begin{table}[t!]
  \setlength\tabcolsep{3.0pt} %2.5pt%
  \caption{Statistics of the Datasets. AR represents the anomaly ratio, calculated as the ratio of the number of anomalies to the total number of nodes.}
  \begin{tabular}{l|cccccc}
    \toprule
    \textbf{Datasets} & \textbf{Nodes} & \textbf{Edges} & \textbf{Attributes} & \textbf{Timestamps}& \textbf{Anomalies} \\
    \midrule
    DBLP-3  & 4,257 & 38,240 & 100 & 10 & 210  \\
    DBLP-5  & 6,606 & 65,915 & 100 & 10 & 330 \\
    Reddit  & 8,291 & 292,400 & 20 & 10 & 420 \\
    Brain  & 5,000 & 1,975,648 & 20 & 12 & 240 \\
    Bitcoin-OTC &6,005  &355,096  &32  & 138 & 300 \\
    DGraph & 3,700,550 & 4,300,999 & 17 & 821 & 15,509 \\
    \bottomrule
  \end{tabular}
  \label{tab:datasets}
\end{table}

Connectivity between two nodes is established if they exhibit a similar level of activation during the observed time frame. BitcoinOTC is a
who-trusts-whom network of bitcoin users trading on
the platforms www.bitcoinotc.com. Nodes represent
the users from the platform, and an edge appears when one user rates another on the platform. DGraph~\cite{huang2022dgraph} is a large-scale financial network, showcasing registered users as nodes, emergency contact relationships as edges, and 17 attributes from users’ personal profiles as node features.

Since only DGraph has ground-truth labels for anomalies, we manually inject synthetic anomalies into the other five datasets for evaluation in the testing phase. For a fair comparison, we follow the anomaly injection strategies used in \cite{ding2019deep} and \cite{liu2021anomaly}, and inject equal numbers of structural anomalies and attributive anomalies for each snapshot $\mathcal{G}^t$ in the test set:
\begin{itemize}
    \item \textbf{Structural anomaly injection.} Following \cite{ding2019deep}, we generate structural anomalies by randomly selecting $N_p$ nodes from node set $\mathcal{V}^t$ and connecting them to form fully connected cliques. The selected $N_p$ nodes are labeled as structural anomaly nodes. This process is repeated $q$ times to generate $q$ cliques. 
    \item \textbf{Attributive anomaly injection.} Following \cite{liu2021anomaly}, attributive anomalies are created by randomly selecting $N_p\times q$ nodes from $\mathcal{V}^t$. For each chosen node $v^t_i$, we sample an additional $k$ nodes to form a candidate set: $\mathcal{V}^t_{i,attr}=\{v^t_1,\ldots,v^t_k\}$. Then, we replace the feature vector of $v_i$ with the node feature from $\mathcal{V}^t_{i,attr}$ that has the largest attribute distance from $v^t_i$. Following \cite{liu2021anomaly}, we set $k$=50 for all the datasets.
\end{itemize}

\begin{table*}[!th]
  \setlength\tabcolsep{9.0pt}
  \renewcommand\arraystretch{0.9}
  \small
  \centering
  \caption{Node anomaly detection performance on six benchmark datasets, the best and second to best results on each dataset are in bold and underlined, respectively. PRE, F1, and AUC represent the precision, macro-F1, and area under the curve, respectively.} 
  \label{tab:node_an}
  \begin{tabular}{l|ccc|ccc|ccc}
    \toprule
    \textbf{Dataset} & \multicolumn{3}{c|}{\textbf{DBLP-3}} & \multicolumn{3}{c|}{\textbf{DBLP-5}} & \multicolumn{3}{c}{\textbf{Reddit}} \\
    \textbf{Metrics} & PRE & F1 & AUC & PRE & F1 & AUC & PRE & F1 & AUC \\  
    \midrule
    DOMINANT &0.1358 &0.5490 &0.6994 &0.5431 &\underline{0.7327} &\underline{0.9154} &0.4912 &0.6326 &\underline{0.9316} \\
    CoLA &0.4753 &0.4874 &0.5814 &0.4750 &0.4872 &0.4806 &0.4747 &0.4870 &0.2410 \\
    SL-GAD &0.5302 &0.5193 &0.6174 & 0.5229 & 0.5038 &0.7638 &0.4908 &0.4844 &0.6843 \\
    GRADATE &0.5126 &0.4874 &0.5581 &0.4923 &0.4875 &0.4726 &0.5231 &0.4870 &0.5823 \\
    \midrule
    NetWalk &0.5336 &0.5092 &0.7126 &0.5805 &0.5034 &0.9107 &0.5606 &0.5726 &0.7821 \\
    MTHL &0.5758 &0.5077 &0.5901 &0.5433 &0.4951 &0.7382 &0.6295 &0.6279 &0.7074  \\
    TADDY &0.4994 &0.5124 &0.6425 &0.4994 &0.5135 &0.6878 &0.5624 &0.5246 &0.7390  \\
    SAD &0.7044 &\underline{0.7207} &0.8973 &0.4931 &0.3523 &0.4535 &0.5198 &0.5219 & 0.8551 \\
    CLDG  &\underline{0.7225} &0.7188 &\underline{0.8882} &\underline{0.7208} &0.7021 &0.8781 &\underline{0.6819} &\underline{0.6601} &0.8348 \\
    \midrule
    \name & \textbf{0.7622} & \textbf{0.7972} &\textbf{0.9620} & \textbf{0.7359} & \textbf{0.8020} & \textbf{0.9765} &\textbf{0.9409} & \textbf{0.6849} & \textbf{0.9810} \\
    % Gain ($\%$) & 66.51 & 61.65 & 34.99 & 57.53  & 24.14 &6.6 &79.87 &8.2  &12.5  &4.2  &7.5  & 7.48  \\
    \bottomrule
    \toprule
\textbf{Dataset} & \multicolumn{3}{c|}{\textbf{Brain}} & \multicolumn{3}{c|}{\textbf{Bitcoin-OTC}} & \multicolumn{3}{c}{\textbf{DGraph}} \\
    \textbf{Metrics} & PRE & F1 & AUC & PRE & F1 & AUC & PRE & F1 & AUC \\  
    \midrule
    DOMINANT & 0.5845 &0.5958 &0.8212 &0.5157 &0.5062 &0.9079 &0.4985 &0.4914 &0.5709 \\
    CoLA &0.4760 &0.4877 &0.5716 &0.4750 &0.4867 &0.7117 &0.3312 &0.3893 &0.4361 \\
    SL-GAD &\underline{0.6640} &0.6347 & \underline{0.8735} &0.4750 &0.4871 &0.9356 & OOM & OOM & OOM \\
    GRADATE &0.4910 & 0.4871&0.5629 &0.4971 &0.4868 &0.7592 &OOM &OOM &OOM \\
    \midrule
    NetWalk &0.6239 &\underline{0.6643} &0.8590 &0.6667 &0.6324 &0.9504 &OOM &OOM &OOM \\
    MTHL  &0.5843 &0.5988 &0.8119 &0.6727 &0.6410 &0.9353 &OOM &OOM &OOM \\
    TADDY &0.6218 &0.6365 &0.7235 &0.6552 &0.6821 &\underline{0.9566} &0.5940 &0.5477 &0.6654 \\
    SAD &0.5266 &0.5065 &0.5638 &0.5733 &0.7277 &0.6341 &0.4248 &\underline{0.6136} &\underline{0.7312}  \\
    CLDG &0.6372 &0.6425 &0.5928 &\underline{0.7315} &\underline{0.7544} &0.8394 &\underline{0.6210} &0.6018 &0.6528 \\
    \midrule
    \name & \textbf{0.6919} & \textbf{0.7144} & \textbf{0.9389} &\textbf{0.7579} &\textbf{0.8259} &\textbf{0.9952} &\textbf{0.6451} &\textbf{0.6514} &\textbf{0.7526} \\
    \bottomrule
\end{tabular}
\end{table*}

\subsection{Baselines}\label{sub:base}
To validate the effectiveness of \name, we conducted a comparative analysis with nine state-of-the-art node anomaly detection baselines. This comparison includes five dynamic node anomaly detection methods: NetWalk~\cite{yu2018netwalk}, MTHL~\cite{teng2017anomaly}, TADDY~\cite{liu2021anomaly2}, SAD~\cite{tian2023sad} and CLDG~\cite{xu2023cldg}. Given the limited number of dynamic node anomaly detection baselines, we also incorporate four of the most advanced static node anomaly detection methods in our comparison: DOMINANT~\cite{ding2019deep}, CoLA~\cite{liu2021anomaly}, SL-GAD~\cite{zheng2021generative} and GRADATE~\cite{duan2023graph}. Details of these baselines are introduced as follows:

\begin{itemize}[leftmargin=*]
    \item[] \textbf{Static node anomaly detection methods:}
\end{itemize}
\begin{itemize}
    \item \textbf{DOMINANT}~\cite{ding2019deep} is a deep graph autoencoder-based unsupervised method that detects node anomalies by assessing the reconstruction errors of individual nodes.
    \item \textbf{CoLA}~\cite{liu2021anomaly} is a contrastive learning based anomaly detection method that captures node anomaly patterns by measuring the agreement between each node and its contextual subgraph using a GNN-based encoder.
    \item \textbf{SL-GAD}~\cite{zheng2021generative} is a self-supervised anomaly detection method that combines both attribute reconstruction and contrastive learning for detecting node anomalies.
    \item  \textbf{GRADATE}~\cite{duan2023graph} is an extension of CoLA by conducting contrastive learning not only between node-node and node-subgraph pairs, but also from subgraph-subgraph pairs.
\end{itemize}
\begin{itemize}[leftmargin=*]
    \item[] \textbf{Dynamic node anomaly detection methods:}
\end{itemize}
\begin{itemize}
    \item \textbf{NetWalk}~\cite{yu2018netwalk} uses an autoencoder to update dynamic node representations and applies streaming k-means clustering for real-time node categorization. Anomaly scores are calculated based on node proximity to cluster centers.
    \item \textbf{MTHL}~\cite{teng2017anomaly} learns a compact hypersphere surrounding normal node representations and then distinguishes normal and abnormal nodes according to their distances to the hypersphere center.
    \item \textbf{TADDY}~\cite{liu2021anomaly2} develops coupled spatial-temporal patterns via a dynamic graph transformer to detect anomalies in graph streams.
    \item \textbf{SAD}~\cite{tian2023sad} is a semi-supervised dynamic graph anomaly detection model that combines memory networks with pseudo-label contrastive learning.
    \item \textbf{CLDG}~\cite{xu2023cldg} is a contrastive learning model that performs representation learning on both discrete-time and continuous-time dynamic graphs.
\end{itemize}

% \subsection{Metrics}\label{sub:metric} 
% To assess the performance of \name and its competitors, we utilize a combination of ROC-AUC, precision, and Macro-F1 as evaluation metrics. ROC-AUC, a standard metric in anomaly detection, involves plotting the true positive rate against the false positive rate, with AUC (Area Under the Curve) representing the area under this ROC curve, ranging between 0 and 1. Higher AUC values indicate better performance. Additionally, we use precision and macro-F1 as our evaluation metrics. Precision measures the proportion of positive identifications that were actually correct. Macro-F1, as the harmonic mean of precision and recall, is computed individually for each class and then averaged, to provide a balanced measure of the model's performance.

\subsection{Parameter Settings}\label{sub:para}
All the parameters can be tuned by 5-fold cross-validation on a rolling basis. We set the time window size $\tau$ as 3 and temporal convolution kernel width $K_t$ as 2. Both the GCN and gated temporal convolution encoders/decoders have 2 layers with hidden dimensions set as 32 for Brain and 128 for the remaining datasets. Balance factor $\alpha$ is set as 0.9 for Brain and Reddit, and 0.3 for DBLP-3, DBLP-5 and OTC. The training epoch is 20 and the learning rate is 0.001 for all the datasets. Evaluation round $R$ is 40 for Brain and 20 for the other datasets. The number of spatial and temporal memory items $P_s$ and $P_t$ are both set as 6. 
For the evaluation of static graph anomaly detection baselines, we first train these methods and measure the anomaly scores for each graph snapshot, then we derive the final score for evaluation by averaging these anomaly scores across all snapshots.

\subsection{Comparison with the State-of-the-art Baselines}\label{sub:detec}
To evaluate the effectiveness of \name on dynamic node anomaly detection, we compare its performance with six state-of-the-art baselines on four benchmark datasets. The precision, macro-F1 and AUC values are presented in Table \ref{tab:node_an}. All the differences between our model and others are statistically significant ($p<0.01$). According to these results, we have the following findings:

\begin{itemize}
    \item The proposed \name consistently outperforms all the baselines on the four dynamic graph datasets, showcasing its superiority in dynamic node anomaly detection. Compared with the most competitive baselines, \name achieves a significant performance gain of 9.5$\%$ on precision, 7.8$\%$ on macro-F1, and 5.8$\%$ on AUC, averagely. This validates the overall design of our proposed \name model.
    \item Compared to static anomaly detection models (DOMINANT, CoLA, SL-GAD, and GRADATE), \name integrates interactions across different timestamps through gated temporal convolution, enabling it to capture the dynamic evolution among graphs. Consequently, \name learns more comprehensive embeddings for dynamic anomaly detection and achieves better performance.
    \item In comparison with most competitive dynamic node anomaly detection methods such as CLDG and SAD, \name achieves 5.8$\%$ average performance gain on AUC. We attribute this performance advantage to our proposed prototype-enhanced reconstruction strategy. Unlike NetWalk\cite{yu2018netwalk} and MTHL\cite{teng2017anomaly} that identify abnormalities based on their distances to cluster centers, our model circumvents the dependence on the selection of clustering techniques and methods for calculating distance. Instead, it directly assesses abnormalities through the reconstruction errors between the original graphs and their reconstructions, which are improved by prototypes. This eliminates potential biases introduced by specific clustering or distance computation methodologies.
\end{itemize} 

\begin{figure}[t!]
  \centering
  \subfigure{
  \includegraphics[scale=0.29]{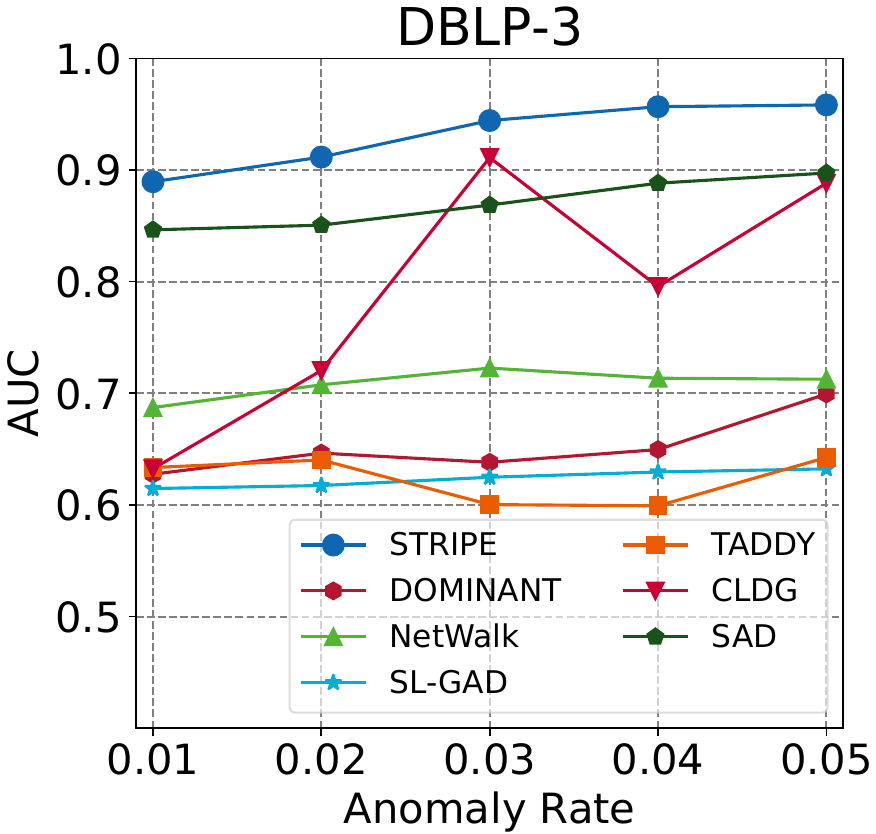}
  }\hspace{-4mm}
  \subfigure{
  \includegraphics[scale=0.29]{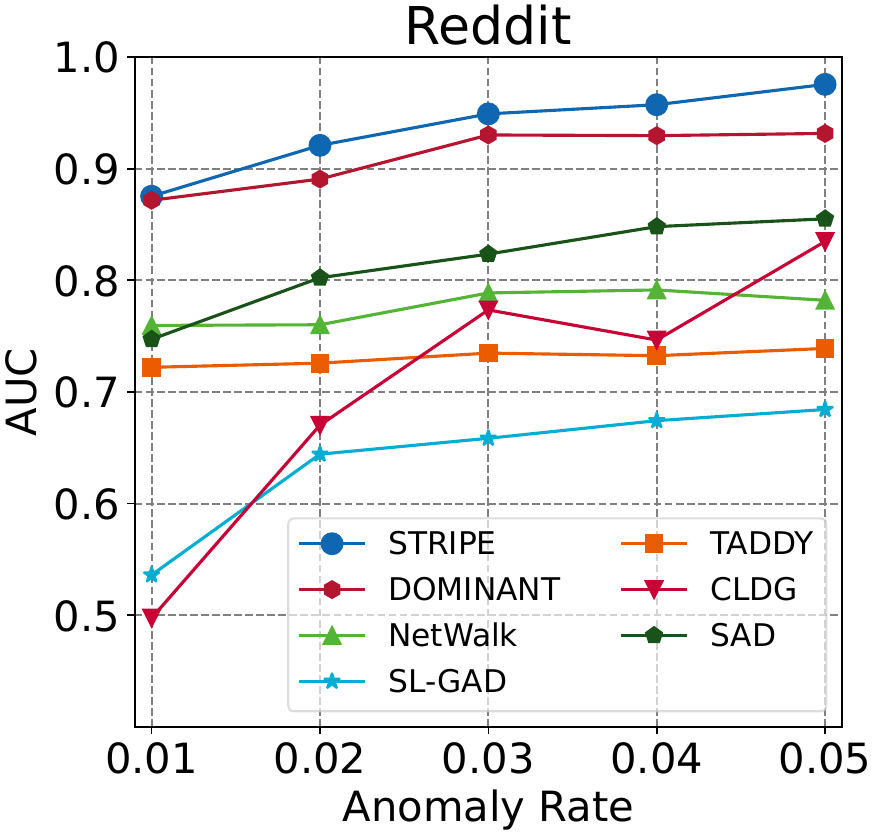}
  }
  \caption{Anomaly detection under different anomaly rates on DBLP-3 (left) and Reddit (right) datasets.} 
  \label{fig:ano_rate}
\end{figure}

\subsection{Performance under Different Anomaly Rates}\label{sub:rate}
To further evaluate the robustness of the proposed model, we further compare the performance of STRIPE with other baselines under different anomaly rates. The anomaly rate is the proportion of anomalous nodes to the total number of nodes, which can be changed by varying the number of injected anomalies $q$ to synthetic datasets. We vary anomaly rate from 0.01 to 0.05 and the result is reported in Fig. \ref{fig:ano_rate}. We can observe that (1) STRIPE consistently outperforms other baselines across all the anomaly rates, which demonstrate the robustness and generalization of our model. (2) STRIPE generally has larger performance gain over other baselines under higher anomaly rates, and only has marginal performance drop compared to baselines such as CLDG or SL-GAD.

\begin{figure}[t!]
  \centering
  \subfigure{
  \includegraphics[scale=0.34]{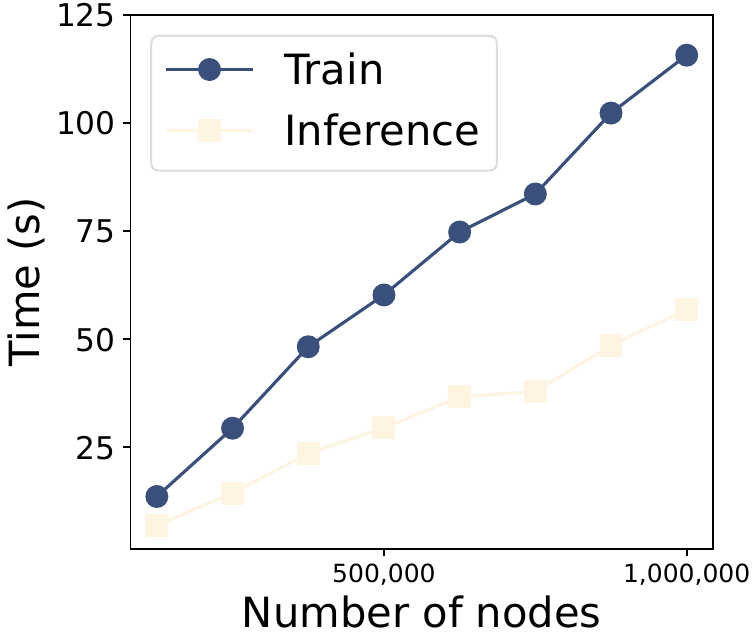}
  }\hspace{-3mm}
  \subfigure{
  \includegraphics[scale=0.34]{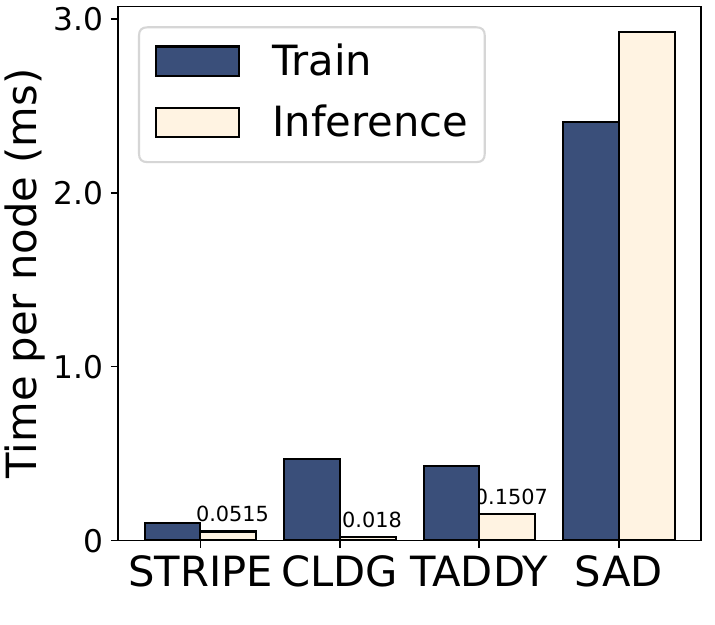}
  }
  \caption{Evaluation of time efficiency. \textbf{Left}: The linear increase of both training and inference time of STRIPE w.r.t. node numbers in DGraph dataset. \textbf{Right}: Comparison of training and inference time of STRIPE with three most competitive baselines on DGraph dataset.} 
  \label{fig:eff}
\end{figure}
\subsection{Model Efficiency Evaluation}\label{sub:eff}
In this subsection, we provide experimental analysis of the computational efficiency of the proposed STRIPE. We measure the running time per epoch of STRIPE on the DGraph dataset while varying the number of nodes. As depicted in the left subplot of Fig. \ref{fig:eff}, both the training and inference time is linear w.r.t. node numbers, which aligns with our analysis in section \ref{sec:complix}. Additionally, we compare the time efficiency of STRIPE with baselines that acquire AUC scores over 60\% on DGraph dataset, as shown in the right subplot of Fig. \ref{fig:eff}. Compared to CLDG, the second most accurate model following STRIPE, despite STRIPE is 2.86$\times$ slower than CLDG during inference, STRIPE is 4.62$\times$ faster in training and demonstrates a significant improvement of 5.8\% in AUC scores compared to CLDG. Furthermore, STRIPE is 23.76$\times$ faster in training and 56.8$\times$ faster in inference compared to the second most accurate baseline, SAD.

\begin{table*}[t!]
  \small
  \centering
  \caption{Quantitative results w.r.t. precision, recall and AUC for ablation study} 
  \label{tab:ablation}
  \begin{tabular}{l|ccc|ccc|ccc|ccc}
    \toprule
    \textbf{Dataset} & \multicolumn{3}{c|}{\textbf{DBLP-3}} & \multicolumn{3}{c|}{\textbf{DBLP-5}} & \multicolumn{3}{c|}{\textbf{Reddit}} & \multicolumn{3}{c}{\textbf{Brain}} \\
    \textbf{Metrics} & PRE & F1 & AUC & PRE & F1 & AUC & PRE & F1 & AUC & PRE & F1 & AUC \\  
    \midrule
    \textit{w/o attribute} &0.7583 &0.7688 &0.9513 &0.7039 &0.7814 &0.9462 &0.5042 &0.5011 &0.8963 &0.5744 &0.5478 &0.7215  \\
    \textit{w/o structure} &0.6756 &0.6739 &0.8351 &0.6581 &0.7990 &0.9721 &0.9344 & 0.6628&0.9603 &0.6674 &0.7046 &0.9263  \\
    \textit{w/o temporary} &0.4996 &0.2371 &0.5145 &0.4949 &0.4849 &0.5547 &0.6914 &0.5879 &0.8081  &0.6919 & 0.6442 & 0.8621 \\
    \textit{w/o s-prototype} &0.7164 &0.7436 &0.9491 &0.7320 &0.7905 &0.9516 &0.7897 & 0.7589 &0.9570 &0.6384 &0.6802 & 0.8480 \\
    \textit{w/o t-prototype} &0.6524 &0.6823 &0.8881 &0.7248 &0.7861 &0.9412 &0.6867  &0.6487 &0.9466 &0.6721 &0.7040 &0.9022 \\
    \midrule
    \name & \textbf{0.7622} & \textbf{0.7972} &\textbf{0.9620} & \textbf{0.7359} & \textbf{0.8020} & \textbf{0.9765} &\textbf{0.9409} & \textbf{0.6849} & \textbf{0.9810} & \textbf{0.6919} & \textbf{0.7144} & \textbf{0.9389} \\
    \bottomrule
\end{tabular}
\end{table*}

\begin{figure*}[t!]
  \centering
  \subfigure[DBLP-3]{
  \includegraphics[scale=0.32]{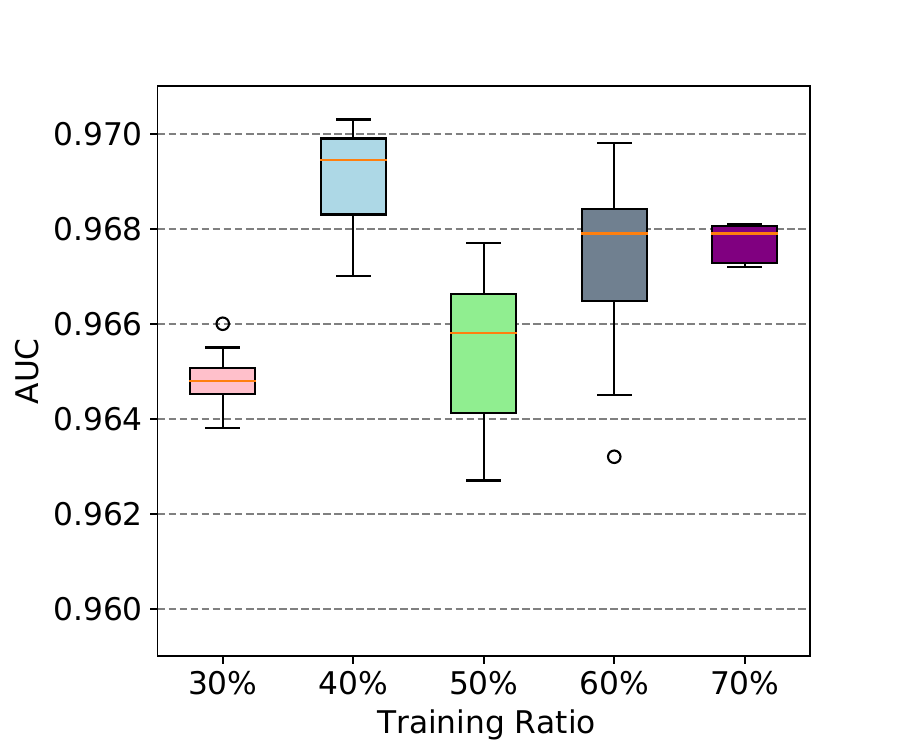}
  }\hspace{-3mm}
  \subfigure[DBLP-5]{
  \includegraphics[scale=0.32]{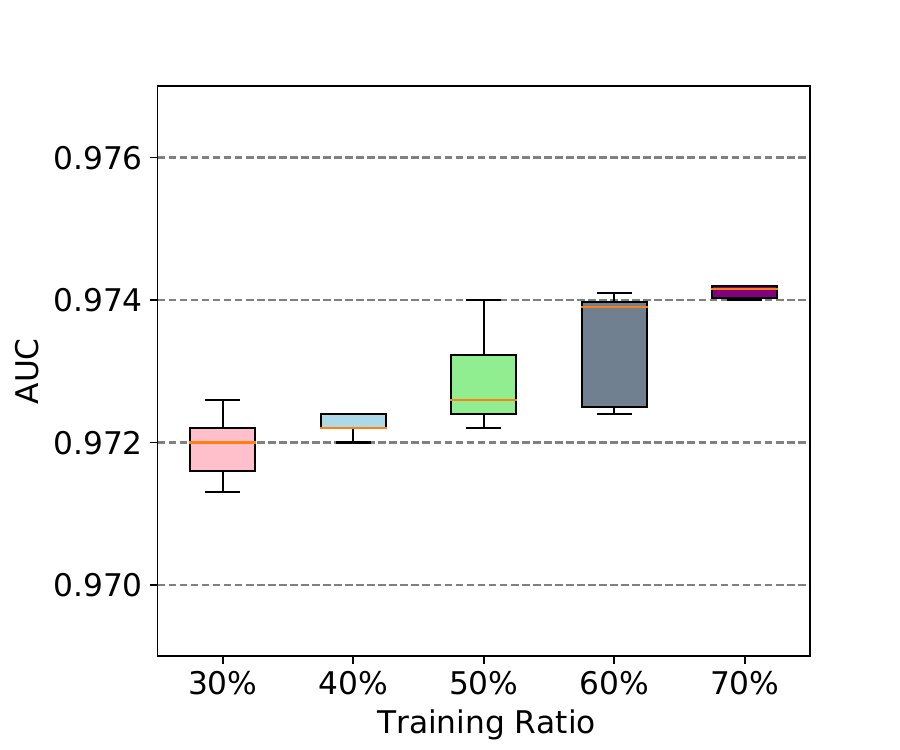}
  }\hspace{-3mm}
  \subfigure[Reddit]{
  \includegraphics[scale=0.32]{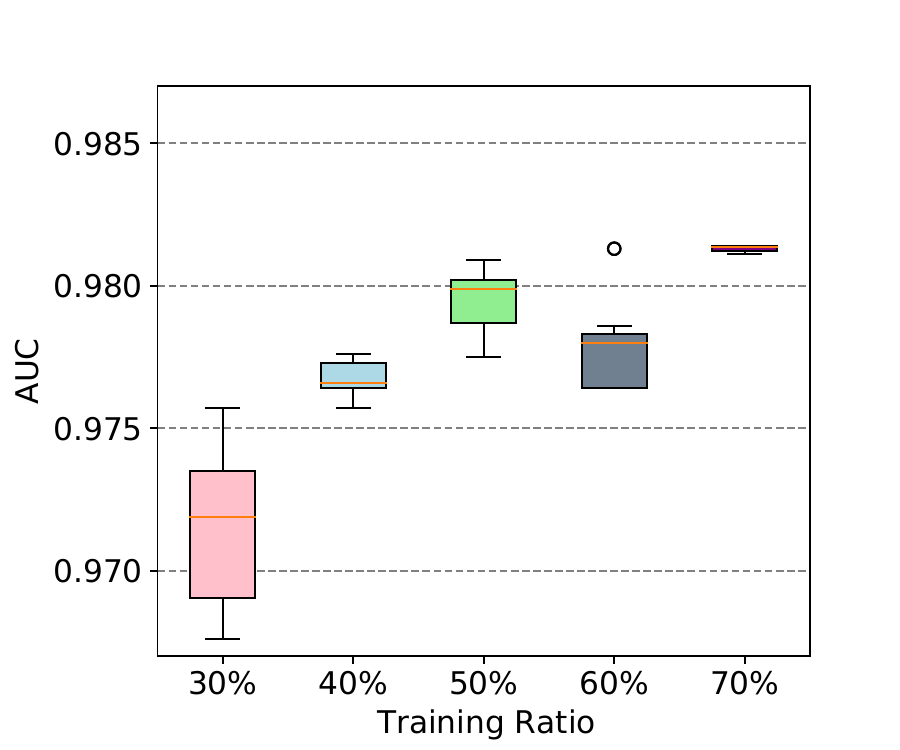}
  }\hspace{-3mm}
  \subfigure[Brain]{
  \includegraphics[scale=0.32]{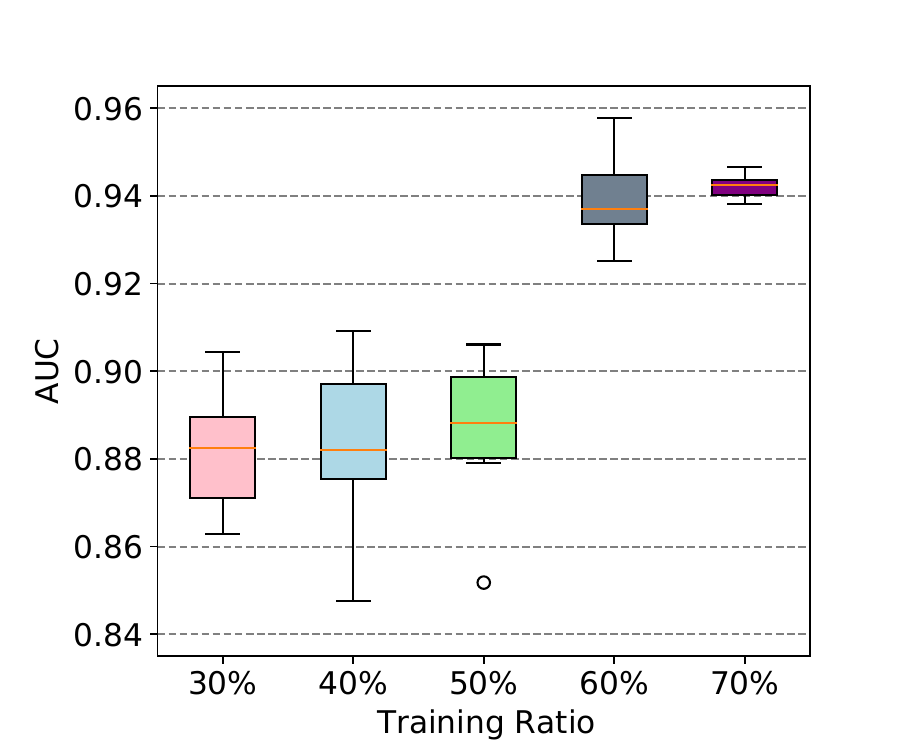}
  }
  \caption{AUC values of \name on four datasets with different training ratios. The circular markers indicate the results which are viewed as outliers.
} 
  \label{fig:train_ratio}
\end{figure*}

\subsection{Ablation Study}\label{sub:ablation}
To further investigate the contribution of each component in \name, we perform an ablation study in this section. The results are shown in Table \ref{tab:ablation}. We set five variants of \name: \textit{w/o attribute}, \textit{w/o structure}, \textit{w/o temporary}, \textit{w/o s-prototype}, \textit{w/o t-prototype} and \name. Among them, \textit{w/o attribute} excludes the attributive reconstruction error by setting $\alpha$=0. \textit{w/o structure} excludes the structural reconstruction error by setting $\alpha$=1. \textit{w/o temporary} excludes the gated temporal convolution component and directly applies static version of \name on each graph snapshot and computes the average anomaly scores across all the snapshots. \textit{w/o s-prototype} and \textit{w/o t-prototype} exclude the spatial and temporal memory modules, respectively, and reconstructs graphs without fusing prototypical patterns.

The comparison between \textit{w/o attribute}, \textit{w/o structure}, and \name, as illustrated in Table \ref{tab:ablation}, reveals a decline in performance when either attributive or structural reconstruction error is excluded from the loss function. This indicates that incorporating both types of reconstruction is essential for optimal performance in the node anomaly detection task. Furthermore, removing the gated temporal convolution component, as shown in the \textit{w/o temporary} comparison with \name, results in the most pronounced decrease in performance. This underscores the significance of tracking graph evolution for dynamic graph anomaly detection. Additionally, the performance difference between \textit{w/o s-prototype} and \name and the performance gap between \textit{w/o t-prototype} and \name further illustrate the contribution of both spatial and temporal normality memory items on the enhancement of anomaly detection performance.

\subsection{Parameter Sensitivity}\label{sub:sens}
In this subsection, we conduct a series of experiments to study the impact of various hyper-parameters in \name, including \textit{training ratio}, \textit{temporal convolution parameters}, \textit{reconstruction weight}, \textit{hidden dimension} and \textit{evaluation rounds}.

\subsubsection{Training ratio}
In this experiment, we assessed the robustness of \name by examining its performance across varying training ratios within the range of $\{30\%, 40\%, 50\%, 60\%, 70\%\}$. As presented in Fig. \ref{fig:train_ratio}, the increase of training ratio results in a general improvement in AUC values across four datasets. This indicates that a larger volume of training data enhances the model's ability to learn the normal pattern in the training set. The performance fluctuation (e.g., 40\% training ratio on the DBLP-3 dataset and 60\% training ratio on the Reddit dataset) is likely due to the shift in data distribution when enlarging the training set. Additionally, a decrease in AUC variance with higher training ratios suggests that \name achieves more stable performance with sufficient training data. Notably, \name maintains competitive performance even at a lower training ratio of $30\%$, demonstrating its robustness to conduct dynamic node anomaly detection from limited training data.

\subsubsection{Number of memory items}\label{subsub:n_item}
\begin{figure*}[t!]
  \centering
  \subfigure[DBLP-3]{
  \includegraphics[scale=0.32]{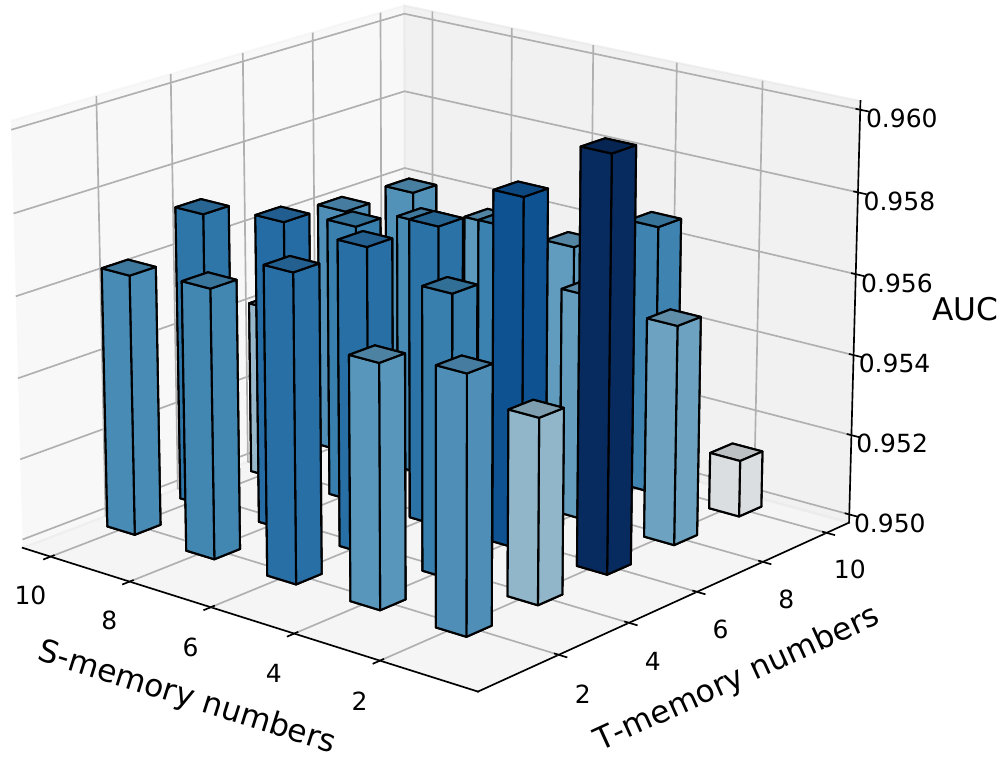}
  }
  \subfigure[Reddit]{
  \includegraphics[scale=0.32]{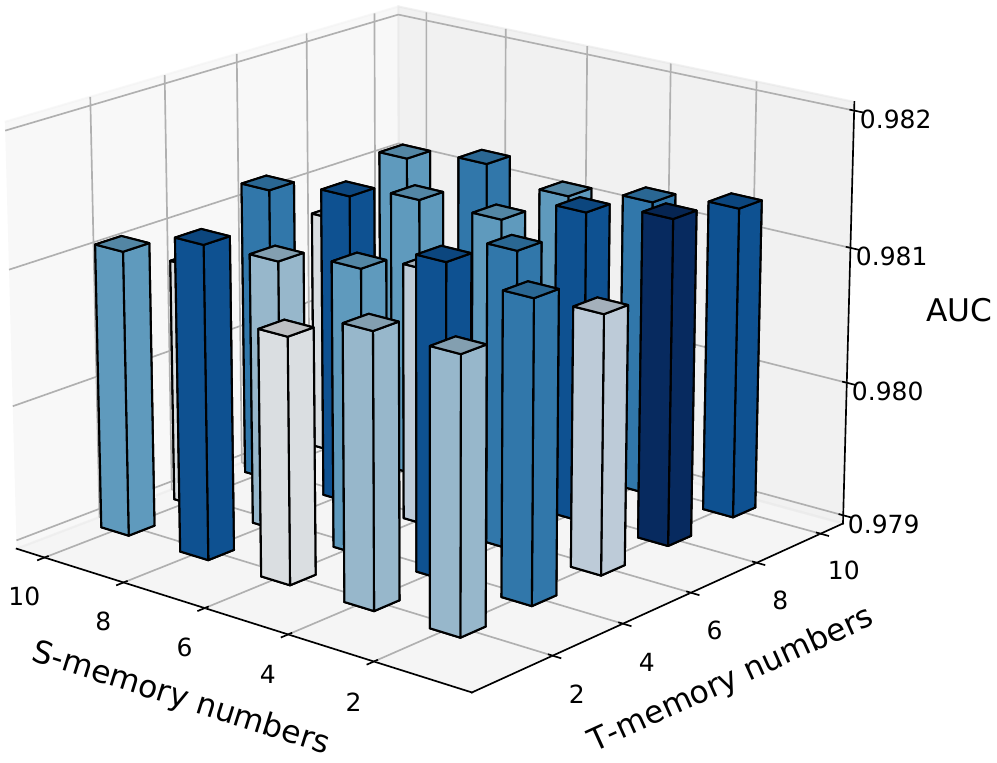}
  }
  \subfigure[Brain]{
  \includegraphics[scale=0.32]{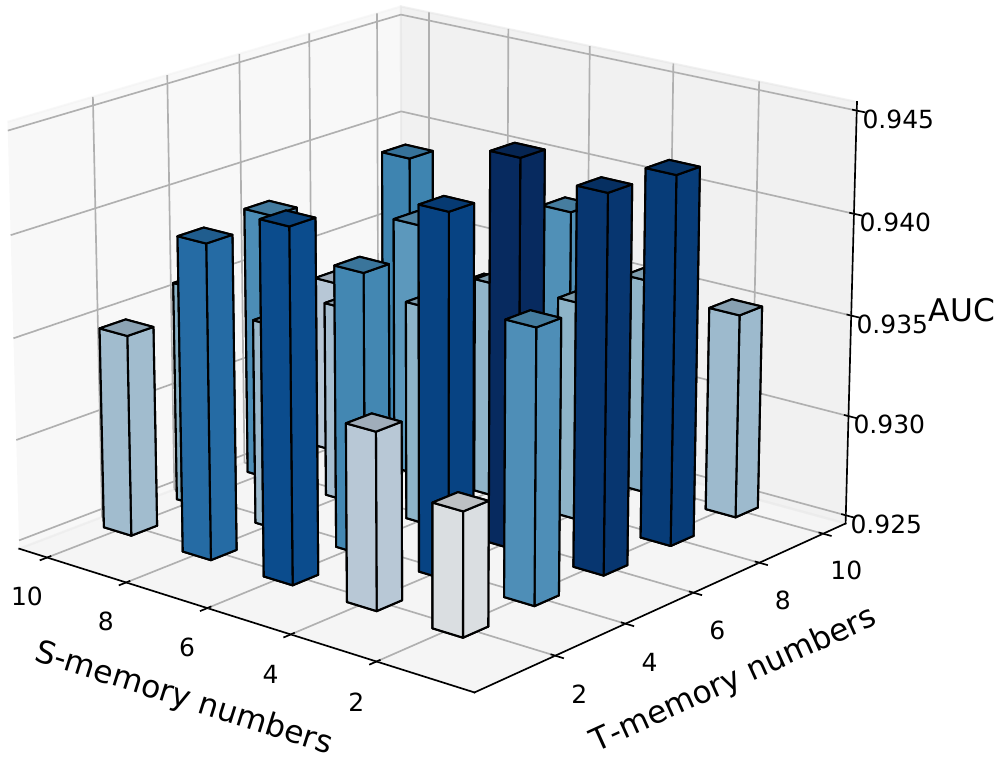}
  }
  \caption{The sensitivity of spatial memory item number $P_s$ and temporal memory item number $P_t$ on three datasets. The vertical axis represents the AUC values of \name with different $P_s$ and $P_t$. A darker color indicates a higher AUC value.
} 
  \label{fig:mem}
\end{figure*}

In this research, we explore the effects of varying the number of spatial memory items ($P_s$) and temporal memory items ($P_t$) within the set $\{2, 4, 6, 8, 10\}$. The sensitivity of our model to $P_s$ and $P_t$ is depicted in Fig. \ref{fig:mem}. From our analysis, we derive three key insights:

The performance of our model, \name, exhibits variability with adjustments in $P_s$ and $P_t$ across the DBLP-3 and Brain datasets. Conversely, the Reddit dataset shows minimal performance fluctuation in response to changes in these parameters, likely due to its simpler normality patterns. This suggests that \name can effectively capture normal spatial and temporal patterns with a minimal number of memory items, maintaining robust anomaly detection performance in simpler datasets.

For the DBLP-3 and Brain datasets, we observe that increasing $P_s$ or $P_t$ from lower values generally leads to higher AUC scores. For example, enhancing $P_t$ from 2 to 6 with $P_s$ fixed at 2 in DBLP-3 improves performance, underscoring that a sparse memory set may not adequately represent the graph's complex patterns. Nonetheless, further increasing the count of memory items beyond a certain point does not always yield better performance; it may in fact impair detection capabilities. For instance, elevating $P_s$ from 6 to 10 while keeping $P_t$ at 2 in Brain illustrates this trend, suggesting that excessively large memory modules might overemphasize specific details over general normality patterns, enabling abnormality to be reconstructed accurately.

The model's performance demonstrates greater sensitivity to variations in $P_t$ compared to $P_s$. This indicates the temporal dimension's intricate prototypical patterns and underscores the importance of separately considering spatial and temporal patterns for more nuanced anomaly detection.

\begin{figure*}[t!]
  \centering
  \subfigure[DBLP-3]{
  \includegraphics[scale=0.32]{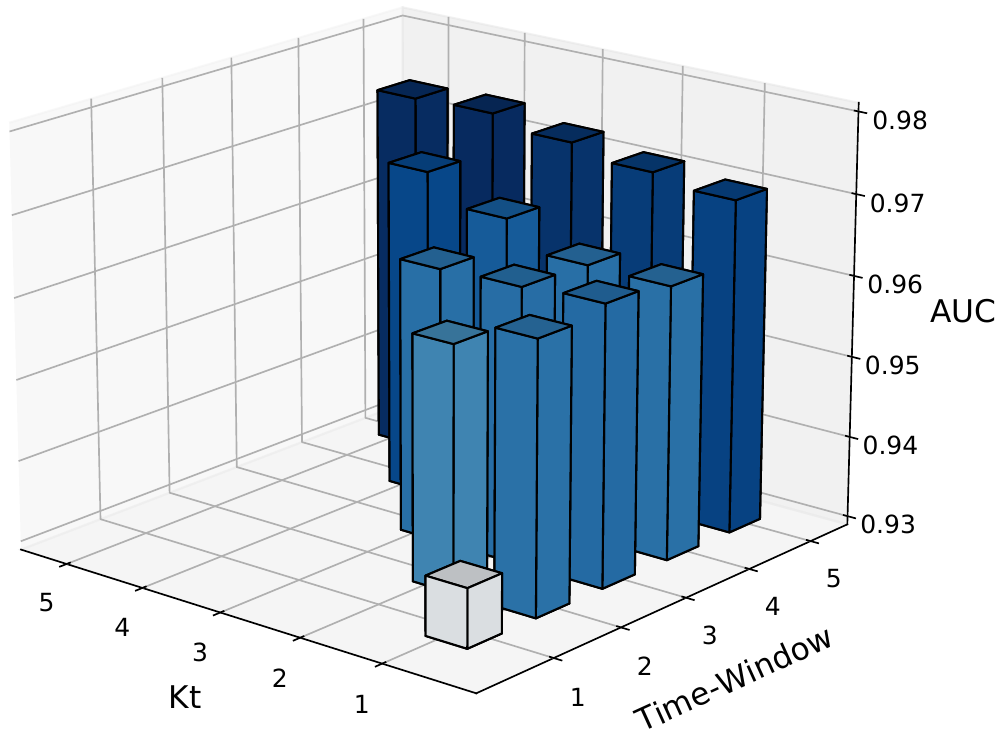}
  }
  \subfigure[Reddit]{
  \includegraphics[scale=0.32]{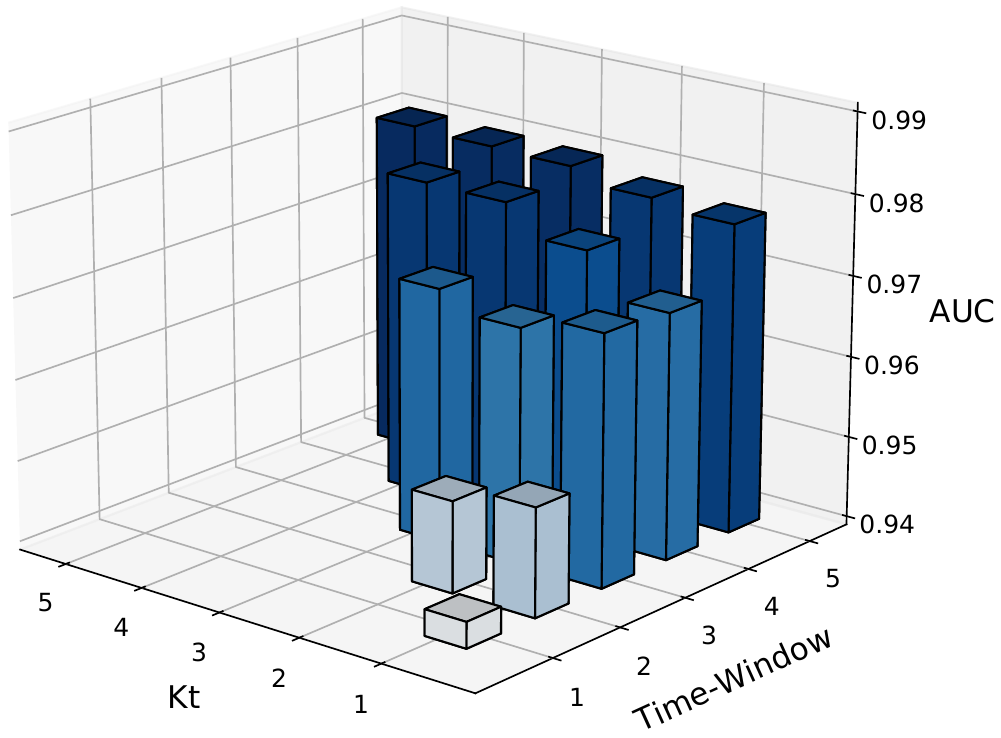}
  }
  \subfigure[Brain]{
  \includegraphics[scale=0.32]{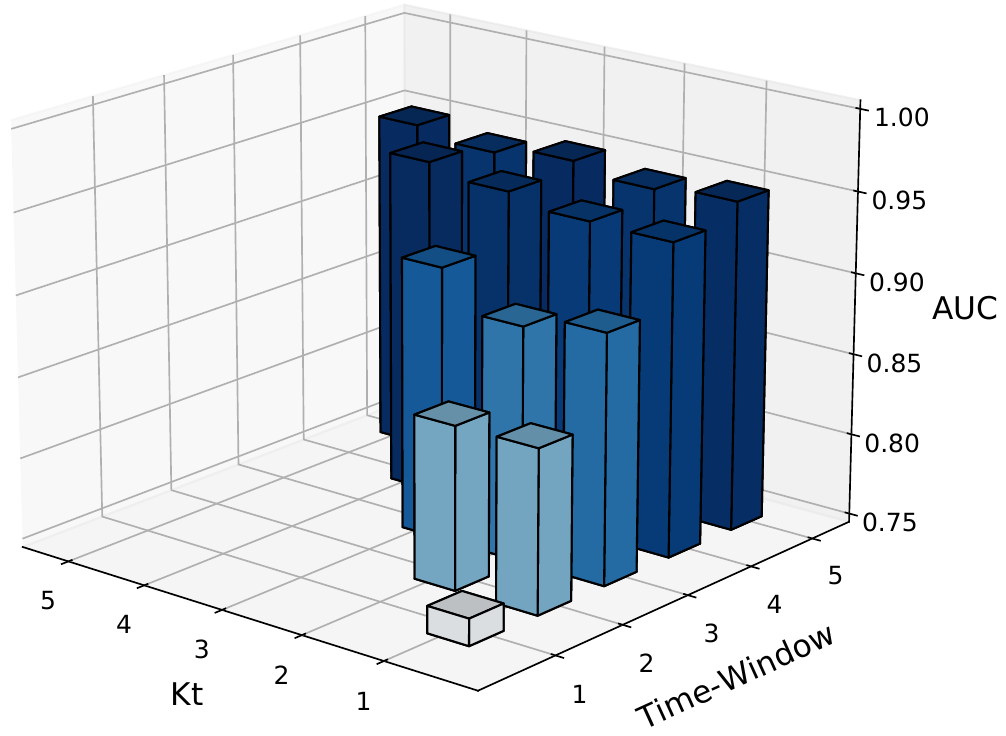}
  }
  \caption{The sensitivity of temporal convolution size $K_t$ and time window size $\tau$ on three datasets. The vertical axis represents the AUC values of \name with different $K_t$ and $\tau$. A darker color indicates a higher AUC value.
} 
  \label{fig:temp_param}
\end{figure*}

\subsubsection{Parameters of temporal convolution}
In this study, we assess the impact of the temporal convolution kernel size ($K_t$) and the number of time window sizes ($\tau$) on temporal convolution performance. We varied $\tau$ across $\{1, 2, 3, 4, 5\}$, ensuring $K_t$ satisfies $1 \leq K_t \leq \tau$ for each value of $\tau$. Sensitivity to changes in $K_t$ and $\tau$ is depicted in Fig. \ref{fig:temp_param}.

The findings reveal suboptimal model performance when the time window size is set to 1, as temporal convolution in this scenario covers only a single graph snapshot, failing to grasp the dynamic evolution across multiple timestamps. As $\tau$ increases, we observe a notable improvement in AUC. detection performance achieves stability when $\tau \geq 4$, suggesting this time window size can adequately capture both short- and long-term dependencies in graph evolution. Further increasing $\tau$ may introduce unnecessary noise and redundancy, detracting from the model performance.

\subsubsection{Reconstruction Weight}
In this experiment, we investigate the influence of the reconstruction weight $\alpha$ in Eq. (\ref{eq:rec_loss}). We vary $\alpha$ from 0.0 to 1.0 and analyze the corresponding AUC values. As depicted in Fig. \ref{fig:para_alpha}. We can observe that for biological and social networks (Brain and Reddit), an increase in $\alpha$ correlates with improved AUC values, peaking at $\alpha$ values of 0.7 and 0.9, respectively. In contrast, for co-authorship networks (DBLP-3 and DBLP-5), optimal AUC values are observed at $\alpha \leq 0.4$, with a performance decline as $\alpha$ increases beyond this point. This observation suggests that the contribution of attributive and structural reconstruction error is affected by the characteristics and domains of different datasets.

\subsubsection{Hidden Dimension}
In this experiment, we explored how varying the hidden dimension $D'$ of both the encoder and decoder affects performance by adjusting $D'$ from 8 to 256 and observing the AUC values. As shown in  \ref{fig:para_hid}, we found that increasing $D'$ enhances model performance within the range of [8, 128] for the DBLP-3, DBLP-5, and Reddit datasets, and within [8, 32] for Brain. Beyond these ranges, performance gains were negligible or even negative. Consequently, we opted for a $D'$ of 32 for Brain and 128 for the other datasets.

\subsubsection{Evaluation Rounds}
\begin{figure*}[t!]
  \centering
  \subfigure[$\alpha$ versus AUC]{
  \includegraphics[scale=0.35]{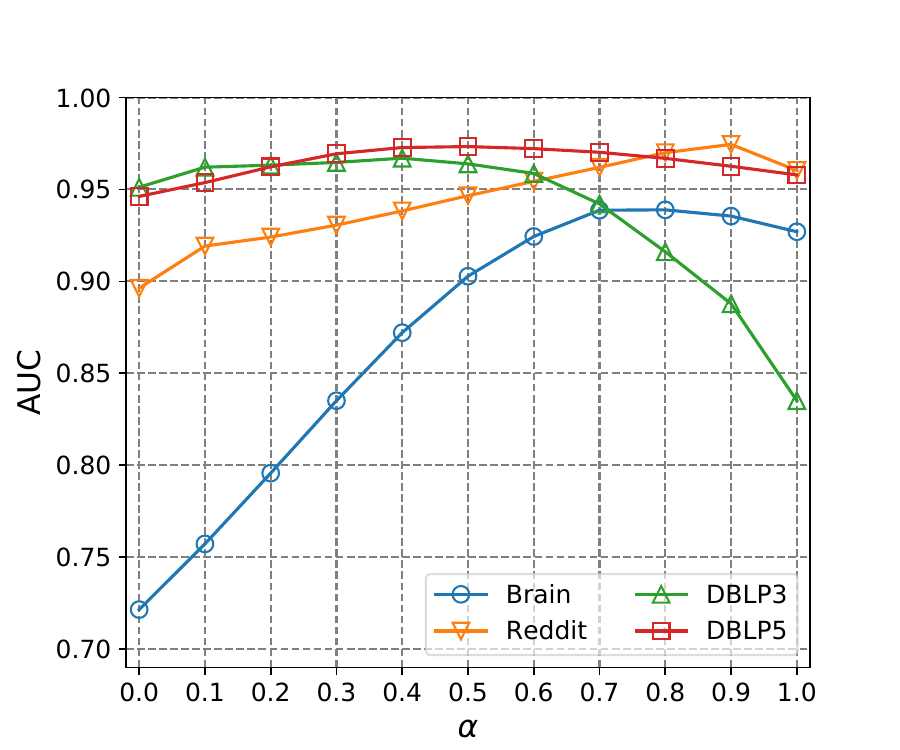}
  \label{fig:para_alpha}}
  \subfigure[Hidden dimension versus AUC]{
  \includegraphics[scale=0.35]{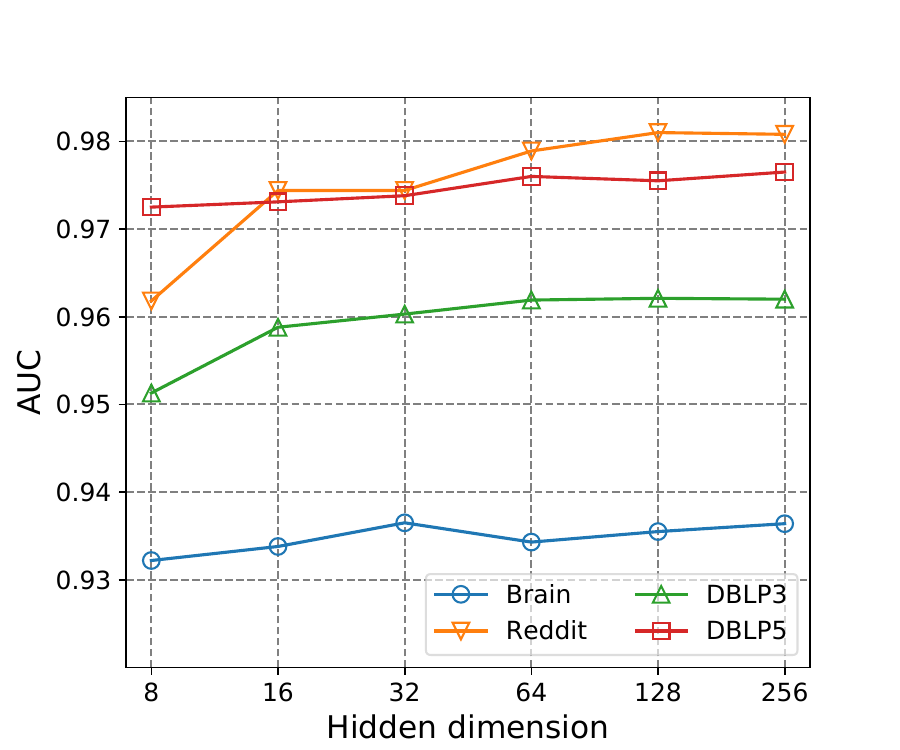}
  \label{fig:para_hid}}
  \subfigure[Evaluation rounds versus AUC]{
  \includegraphics[scale=0.35]{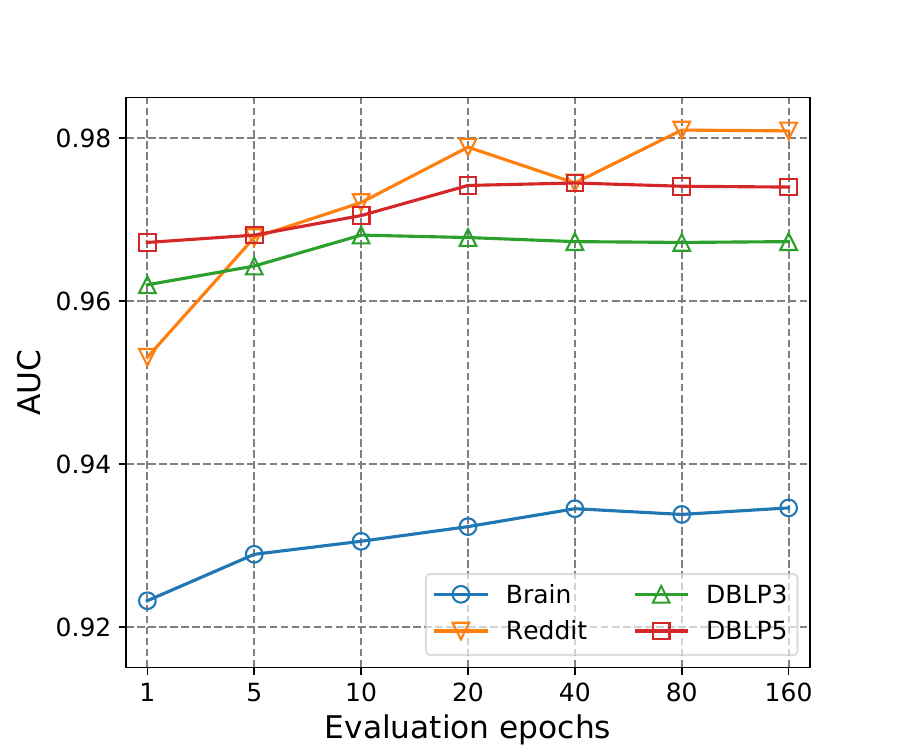}
  \label{fig:para_evl}}
  \caption{AUC value of \name on DBLP3, DBLP5, Reddit and Brain w.r.t. weight $\alpha$, hidden dimension $D'$ and evaluation rounds $R$.
} 
  \label{fig:param}
\end{figure*}

In this section, we evaluated the sensitivity of \name to the number of evaluation rounds (\textit{R}). We adjust \textit{R} from 1 to 160 to observe its impact and depict the results in Fig. \ref{fig:para_evl}. The results indicate poor detection performance at \textit{R}=1, suggesting that a minimal number of rounds fails to adequately detect node anomalies. Performance increases for all the datasets with an increase in \textit{R}. However, elevating \textit{R} beyond 40 for the Brain dataset and 20 for the others does not significantly enhance results but does lead to increased computational demands. Consequently, to optimize both performance and efficiency, we establish \textit{R} at 40 for the Brain dataset and 20 for the remaining datasets.

\begin{figure}[t!]
  \centering
  \subfigure[The compactness loss of normal and abnormal nodes w.r.t. spatial memories.]{
  \includegraphics[scale=0.38]{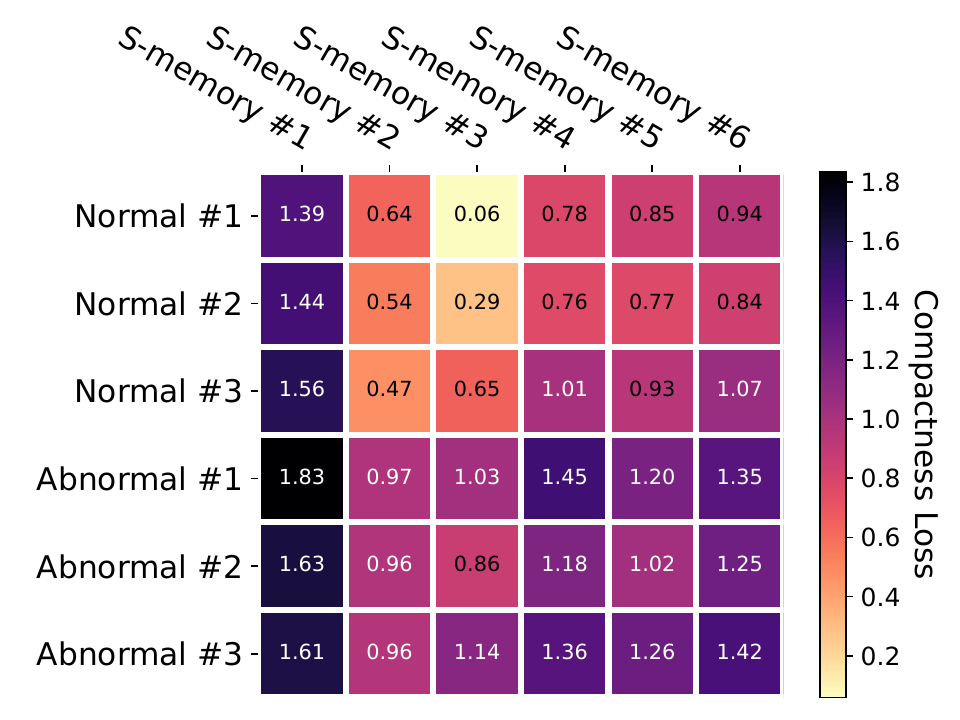}
  \label{fig:s-case}}
  \subfigure[The compactness loss of normal and abnormal nodes w.r.t. temporal memories.]{
  \includegraphics[scale=0.38]{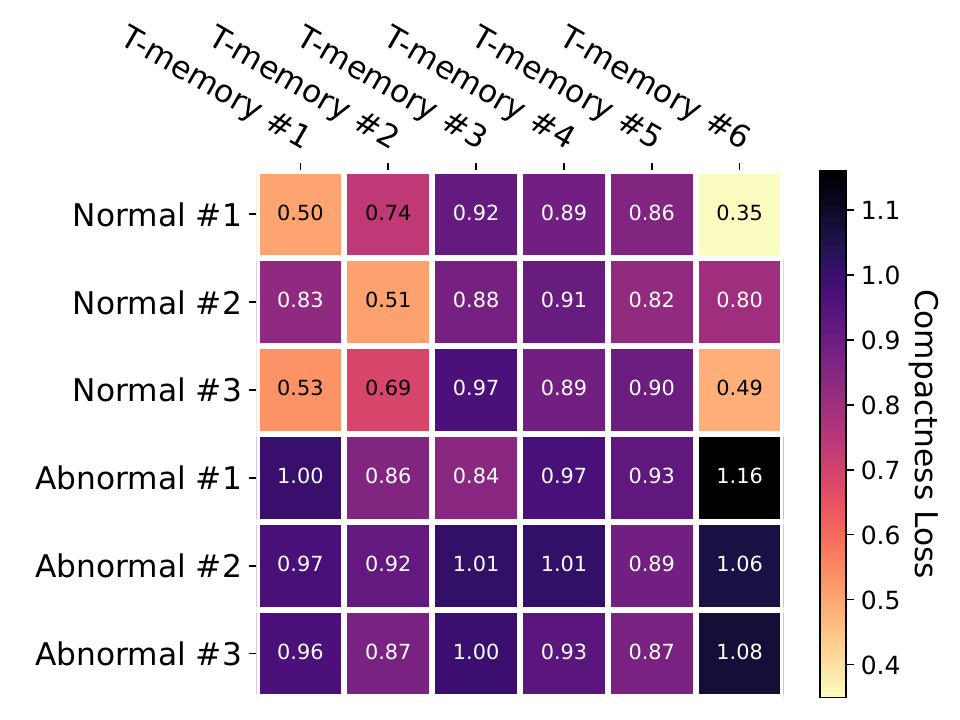}
  \label{fig:t-case}}
  \caption{The compactness loss of normal and abnormal nodes on DBLP-3 with respect to learned spatial and temporal memory items, respectively. A darker color indicates a larger compactness loss between the node feature and the memory.
} 
  \label{fig:case}
\end{figure}

\subsection{Case Study}\label{sub:case}
In this section, we demonstrate the effectiveness of our spatial and temporal memory modules by visualizing the compactness loss between memory items with both normal and abnormal nodes. We configured the spatial and temporal memory items to six each and chose three normal and three abnormal nodes from the DBLP-3 test set for analysis. The compactness loss is calculated as outlined in Eq. \ref{eq:com_loss}. The findings are presented in Fig. \ref{fig:case}, with darker colors indicating higher compactness loss and vice versa. 

From Figs. \ref{fig:s-case} and \ref{fig:t-case}, we note two key observations: (1) Normal nodes exhibit proximity to only a subset of memory items, showing distance from the rest. For instance, Fig. \ref{fig:s-case} illustrates that the compactness loss between the first normal node and the third spatial memory item is merely 0.06, considerably less than its loss with other memory items. This underscores the memory items' capacity to distinctively and effectively encapsulate normal patterns, even with a limited set of items, while ensuring sufficient separation among them. (2) Conversely, abnormal nodes display substantial compactness losses across all memory items in both spatial and temporal dimensions. As an example, the compactness losses between the second abnormal node and all memory items exceed 0.89, highlighting the anomaly nodes' divergence from established normal patterns. This deviation contributes to increased reconstruction errors and, consequently, elevated anomaly scores, effectively signaling their anomalous nature.

\section{Conclusion}
In this study, we addressed the intricate challenge of anomaly detection in dynamic graphs, a domain characterized by the evolving nature of network structures and node attributes. Recognizing the limitations inherent in existing unsupervised learning frameworks, which may struggle to accurately identify anomalies due to their reliance on indirect proxy tasks or their failure to distinguish between spatial and temporal patterns, we proposed a spatial-temporal memories enhanced graph autoencoder (STRIPE) framework. STRIPE represents a novel and comprehensive approach to dynamic graph anomaly detection, distinguished by its meticulous separation and integration of spatial and temporal normality patterns. Extensive evaluation demonstrates that \name significantly outperforms existing methodologies.

\section*{Acknowledgment}
This work is partially supported by the National Key Research and Development Program of China (Grant No. 2020AAA0108504).

\bibliographystyle{IEEEtran}

\bibliography{main}

\begin{thebibliography}{10}
\providecommand{\url}[1]{#1}
\csname url@samestyle\endcsname
\providecommand{\newblock}{\relax}
\providecommand{\bibinfo}[2]{#2}
\providecommand{\BIBentrySTDinterwordspacing}{\spaceskip=0pt\relax}
\providecommand{\BIBentryALTinterwordstretchfactor}{4}
\providecommand{\BIBentryALTinterwordspacing}{\spaceskip=\fontdimen2\font plus
\BIBentryALTinterwordstretchfactor\fontdimen3\font minus \fontdimen4\font\relax}
\providecommand{\BIBforeignlanguage}[2]{{%
\expandafter\ifx\csname l@#1\endcsname\relax
\typeout{** WARNING: IEEEtran.bst: No hyphenation pattern has been}%
\typeout{** loaded for the language `#1'. Using the pattern for}%
\typeout{** the default language instead.}%
\else
\language=\csname l@#1\endcsname
\fi
#2}}
\providecommand{\BIBdecl}{\relax}
\BIBdecl

\bibitem{han2020traffic}
X.~Han, T.~Grubenmann, R.~Cheng, S.~C. Wong, X.~Li, and W.~Sun, ``Traffic incident detection: A trajectory-based approach,'' in \emph{2020 IEEE 36th International Conference on Data Engineering (ICDE)}.\hskip 1em plus 0.5em minus 0.4em\relax IEEE, 2020, pp. 1866--1869.

\bibitem{zheng2019addgraph}
L.~Zheng, Z.~Li, J.~Li, Z.~Li, and J.~Gao, ``Addgraph: Anomaly detection in dynamic graph using attention-based temporal gcn.'' in \emph{IJCAI}, vol.~3, 2019, p.~7.

\bibitem{zhang2021double}
J.~Zhang, M.~Gao, J.~Yu, L.~Guo, J.~Li, and H.~Yin, ``Double-scale self-supervised hypergraph learning for group recommendation,'' in \emph{Proceedings of the 30th ACM international conference on information \& knowledge management}, 2021, pp. 2557--2567.

\bibitem{zheng2016keyword}
B.~Zheng, K.~Zheng, X.~Xiao, H.~Su, H.~Yin, X.~Zhou, and G.~Li, ``Keyword-aware continuous knn query on road networks,'' in \emph{2016 IEEE 32Nd international conference on data engineering (ICDE)}.\hskip 1em plus 0.5em minus 0.4em\relax IEEE, 2016, pp. 871--882.

\bibitem{ranshous2016scalable}
S.~Ranshous, S.~Harenberg, K.~Sharma, and N.~F. Samatova, ``A scalable approach for outlier detection in edge streams using sketch-based approximations,'' in \emph{Proceedings of the 2016 SIAM international conference on data mining}.\hskip 1em plus 0.5em minus 0.4em\relax SIAM, 2016, pp. 189--197.

\bibitem{yu2018netwalk}
W.~Yu, W.~Cheng, C.~C. Aggarwal, K.~Zhang, H.~Chen, and W.~Wang, ``Netwalk: A flexible deep embedding approach for anomaly detection in dynamic networks,'' in \emph{Proceedings of the 24th ACM SIGKDD international conference on knowledge discovery \& data mining}, 2018, pp. 2672--2681.

\bibitem{yang2023time}
Y.~Yang, H.~Yin, J.~Cao, T.~Chen, Q.~V.~H. Nguyen, X.~Zhou, and L.~Chen, ``Time-aware dynamic graph embedding for asynchronous structural evolution,'' \emph{IEEE Transactions on Knowledge and Data Engineering}, 2023.

\bibitem{ma2021comprehensive}
X.~Ma, J.~Wu, S.~Xue, J.~Yang, C.~Zhou, Q.~Z. Sheng, H.~Xiong, and L.~Akoglu, ``A comprehensive survey on graph anomaly detection with deep learning,'' \emph{IEEE Transactions on Knowledge and Data Engineering}, 2021.

\bibitem{wang2021decoupling}
Y.~Wang, J.~Zhang, S.~Guo, H.~Yin, C.~Li, and H.~Chen, ``Decoupling representation learning and classification for gnn-based anomaly detection,'' in \emph{Proceedings of the 44th international ACM SIGIR conference on research and development in information retrieval}, 2021, pp. 1239--1248.

\bibitem{liu2023bourne}
J.~Liu, M.~He, X.~Shang, J.~Shi, B.~Cui, and H.~Yin, ``Bourne: Bootstrapped self-supervised learning framework for unified graph anomaly detection,'' \emph{arXiv preprint arXiv:2307.15244}, 2023.

\bibitem{liu2021anomaly}
Y.~Liu, Z.~Li, S.~Pan, C.~Gong, C.~Zhou, and G.~Karypis, ``Anomaly detection on attributed networks via contrastive self-supervised learning,'' \emph{IEEE transactions on neural networks and learning systems}, vol.~33, no.~6, pp. 2378--2392, 2021.

\bibitem{teng2017anomaly}
X.~Teng, Y.-R. Lin, and X.~Wen, ``Anomaly detection in dynamic networks using multi-view time-series hypersphere learning,'' in \emph{Proceedings of the 2017 ACM on Conference on Information and Knowledge Management}, 2017, pp. 827--836.

\bibitem{zheng2019one}
P.~Zheng, S.~Yuan, X.~Wu, J.~Li, and A.~Lu, ``One-class adversarial nets for fraud detection,'' in \emph{Proceedings of the AAAI Conference on Artificial Intelligence}, vol.~33, no.~01, 2019, pp. 1286--1293.

\bibitem{cai2021structural}
L.~Cai, Z.~Chen, C.~Luo, J.~Gui, J.~Ni, D.~Li, and H.~Chen, ``Structural temporal graph neural networks for anomaly detection in dynamic graphs,'' in \emph{Proceedings of the 30th ACM international conference on Information \& Knowledge Management}, 2021, pp. 3747--3756.

\bibitem{tian2023sad}
S.~Tian, J.~Dong, J.~Li, W.~Zhao, X.~Xu, B.~Song, C.~Meng, T.~Zhang, L.~Chen \emph{et~al.}, ``Sad: Semi-supervised anomaly detection on dynamic graphs,'' \emph{arXiv preprint arXiv:2305.13573}, 2023.

\bibitem{xu2023cldg}
Y.~Xu, B.~Shi, T.~Ma, B.~Dong, H.~Zhou, and Q.~Zheng, ``Cldg: Contrastive learning on dynamic graphs,'' in \emph{2023 IEEE 39th International Conference on Data Engineering (ICDE)}.\hskip 1em plus 0.5em minus 0.4em\relax IEEE, 2023, pp. 696--707.

\bibitem{weston2014memory}
J.~Weston, S.~Chopra, and A.~Bordes, ``Memory networks,'' \emph{arXiv preprint arXiv:1410.3916}, 2014.

\bibitem{huang2023unsupervised}
Y.~Huang, L.~Wang, F.~Zhang, and X.~Lin, ``Unsupervised graph outlier detection: Problem revisit, new insight, and superior method,'' in \emph{2023 IEEE 39th International Conference on Data Engineering (ICDE)}.\hskip 1em plus 0.5em minus 0.4em\relax IEEE, 2023, pp. 2565--2578.

\bibitem{li2017radar}
J.~Li, H.~Dani, X.~Hu, and H.~Liu, ``Radar: Residual analysis for anomaly detection in attributed networks.'' in \emph{IJCAI}, vol.~17, 2017, pp. 2152--2158.

\bibitem{liu2017accelerated}
N.~Liu, X.~Huang, and X.~Hu, ``Accelerated local anomaly detection via resolving attributed networks.'' in \emph{IJCAI}, 2017, pp. 2337--2343.

\bibitem{peng2018anomalous}
Z.~Peng, M.~Luo, J.~Li, H.~Liu, Q.~Zheng \emph{et~al.}, ``Anomalous: A joint modeling approach for anomaly detection on attributed networks.'' in \emph{IJCAI}, 2018, pp. 3513--3519.

\bibitem{ding2019deep}
K.~Ding, J.~Li, R.~Bhanushali, and H.~Liu, ``Deep anomaly detection on attributed networks,'' in \emph{Proceedings of the 2019 SIAM International Conference on Data Mining}.\hskip 1em plus 0.5em minus 0.4em\relax SIAM, 2019, pp. 594--602.

\bibitem{he2024ada}
J.~He, Q.~Xu, Y.~Jiang, Z.~Wang, and Q.~Huang, ``Ada-gad: Anomaly-denoised autoencoders for graph anomaly detection,'' in \emph{Proceedings of the AAAI Conference on Artificial Intelligence}, vol.~38, no.~8, 2024, pp. 8481--8489.

\bibitem{zheng2021generative}
Y.~Zheng, M.~Jin, Y.~Liu, L.~Chi, K.~T. Phan, and Y.-P.~P. Chen, ``Generative and contrastive self-supervised learning for graph anomaly detection,'' \emph{IEEE Transactions on Knowledge and Data Engineering}, 2021.

\bibitem{xu2022contrastive}
Z.~Xu, X.~Huang, Y.~Zhao, Y.~Dong, and J.~Li, ``Contrastive attributed network anomaly detection with data augmentation,'' in \emph{Pacific-Asia conference on knowledge discovery and data mining}.\hskip 1em plus 0.5em minus 0.4em\relax Springer, 2022, pp. 444--457.

\bibitem{ji2013incremental}
T.~Ji, D.~Yang, and J.~Gao, ``Incremental local evolutionary outlier detection for dynamic social networks,'' in \emph{Machine Learning and Knowledge Discovery in Databases: European Conference, ECML PKDD 2013, Prague, Czech Republic, September 23-27, 2013, Proceedings, Part II 13}.\hskip 1em plus 0.5em minus 0.4em\relax Springer, 2013, pp. 1--15.

\bibitem{liu2023meta}
J.~Liu, L.~Song, G.~Wang, and X.~Shang, ``Meta-hgt: Metapath-aware hypergraph transformer for heterogeneous information network embedding,'' \emph{Neural Networks}, vol. 157, pp. 65--76, 2023.

\bibitem{liu2023imbalanced}
J.~Liu, M.~He, G.~Wang, N.~Q.~V. Hung, X.~Shang, and H.~Yin, ``Imbalanced node classification beyond homophilic assumption,'' \emph{arXiv preprint arXiv:2304.14635}, 2023.

\bibitem{aggarwal2011outlier}
C.~C. Aggarwal, Y.~Zhao, and S.~Y. Philip, ``Outlier detection in graph streams,'' in \emph{2011 IEEE 27th international conference on data engineering}, IEEE.\hskip 1em plus 0.5em minus 0.4em\relax IEEE, 2011, pp. 399--409.

\bibitem{sricharan2014localizing}
K.~Sricharan and K.~Das, ``Localizing anomalous changes in time-evolving graphs,'' in \emph{Proceedings of the 2014 ACM SIGMOD international conference on Management of data}, 2014, pp. 1347--1358.

\bibitem{manzoor2016fast}
E.~Manzoor, S.~M. Milajerdi, and L.~Akoglu, ``Fast memory-efficient anomaly detection in streaming heterogeneous graphs,'' in \emph{Proceedings of the 22nd ACM SIGKDD International Conference on Knowledge Discovery and Data Mining}, 2016, pp. 1035--1044.

\bibitem{chen2012community}
Z.~Chen, W.~Hendrix, and N.~F. Samatova, ``Community-based anomaly detection in evolutionary networks,'' \emph{Journal of Intelligent Information Systems}, vol.~39, no.~1, pp. 59--85, 2012.

\bibitem{eswaran2018spotlight}
D.~Eswaran, C.~Faloutsos, S.~Guha, and N.~Mishra, ``Spotlight: Detecting anomalies in streaming graphs,'' in \emph{Proceedings of the 24th ACM SIGKDD International Conference on Knowledge Discovery \& Data Mining}, 2018, pp. 1378--1386.

\bibitem{liu2021anomaly2}
Y.~Liu, S.~Pan, Y.~G. Wang, F.~Xiong, L.~Wang, Q.~Chen, and V.~C. Lee, ``Anomaly detection in dynamic graphs via transformer,'' \emph{IEEE Transactions on Knowledge and Data Engineering}, vol.~35, no.~12, pp. 12\,081--12\,094, 2021.

\bibitem{sukhbaatar2015end}
S.~Sukhbaatar, J.~Weston, R.~Fergus \emph{et~al.}, ``End-to-end memory networks,'' \emph{Advances in neural information processing systems}, vol.~28, 2015.

\bibitem{kim2018memorization}
Y.~Kim, M.~Kim, and G.~Kim, ``Memorization precedes generation: Learning unsupervised gans with memory networks,'' in \emph{International Conference on Learning Representations}, 2018.

\bibitem{wu2018unsupervised}
Z.~Wu, Y.~Xiong, S.~X. Yu, and D.~Lin, ``Unsupervised feature learning via non-parametric instance discrimination,'' in \emph{Proceedings of the IEEE conference on computer vision and pattern recognition}, 2018, pp. 3733--3742.

\bibitem{santoro2016meta}
A.~Santoro, S.~Bartunov, M.~Botvinick, D.~Wierstra, and T.~Lillicrap, ``Meta-learning with memory-augmented neural networks,'' in \emph{International conference on machine learning}.\hskip 1em plus 0.5em minus 0.4em\relax PMLR, 2016, pp. 1842--1850.

\bibitem{cai2018memory}
Q.~Cai, Y.~Pan, T.~Yao, C.~Yan, and T.~Mei, ``Memory matching networks for one-shot image recognition,'' in \emph{Proceedings of the IEEE conference on computer vision and pattern recognition}, 2018, pp. 4080--4088.

\bibitem{zhu2019dm}
M.~Zhu, P.~Pan, W.~Chen, and Y.~Yang, ``Dm-gan: Dynamic memory generative adversarial networks for text-to-image synthesis,'' in \emph{Proceedings of the IEEE/CVF conference on computer vision and pattern recognition}, 2019, pp. 5802--5810.

\bibitem{bengio2006greedy}
Y.~Bengio, P.~Lamblin, D.~Popovici, and H.~Larochelle, ``Greedy layer-wise training of deep networks,'' \emph{Advances in neural information processing systems}, vol.~19, 2006.

\bibitem{kipf2016variational}
T.~N. Kipf and M.~Welling, ``Variational graph auto-encoders,'' \emph{arXiv preprint arXiv:1611.07308}, 2016.

\bibitem{gong2019memorizing}
D.~Gong, L.~Liu, V.~Le, B.~Saha, M.~R. Mansour, S.~Venkatesh, and A.~v.~d. Hengel, ``Memorizing normality to detect anomaly: Memory-augmented deep autoencoder for unsupervised anomaly detection,'' in \emph{Proceedings of the IEEE/CVF International Conference on Computer Vision}, 2019, pp. 1705--1714.

\bibitem{park2020learning}
H.~Park, J.~Noh, and B.~Ham, ``Learning memory-guided normality for anomaly detection,'' in \emph{Proceedings of the IEEE/CVF conference on computer vision and pattern recognition}, 2020, pp. 14\,372--14\,381.

\bibitem{niu2023graph}
C.~Niu, G.~Pang, and L.~Chen, ``Graph-level anomaly detection via hierarchical memory networks,'' in \emph{Joint European Conference on Machine Learning and Knowledge Discovery in Databases}.\hskip 1em plus 0.5em minus 0.4em\relax Springer, 2023, pp. 201--218.

\bibitem{liu2022learning}
Y.~Liu, J.~Liu, M.~Zhao, D.~Yang, X.~Zhu, and L.~Song, ``Learning appearance-motion normality for video anomaly detection,'' in \emph{2022 IEEE International Conference on Multimedia and Expo (ICME)}.\hskip 1em plus 0.5em minus 0.4em\relax IEEE, 2022, pp. 1--6.

\bibitem{huang2022dgraph}
X.~Huang, Y.~Yang, Y.~Wang, C.~Wang, Z.~Zhang, J.~Xu, and L.~Chen, ``{DG}raph: A large-scale financial dataset for graph anomaly detection,'' in \emph{Thirty-sixth Conference on Neural Information Processing Systems Datasets and Benchmarks Track}, 2022.

\bibitem{duan2023graph}
J.~Duan, S.~Wang, P.~Zhang, E.~Zhu, J.~Hu, H.~Jin, Y.~Liu, and Z.~Dong, ``Graph anomaly detection via multi-scale contrastive learning networks with augmented view,'' in \emph{Proceedings of the AAAI Conference on Artificial Intelligence}, vol.~37, no.~6, 2023, pp. 7459--7467.

\end{thebibliography}

\end{document}